\documentclass{article}
\usepackage{iclr2027_conference,times}

\usepackage{microtype}
\usepackage{array}
\usepackage{booktabs}
\usepackage{longtable}   
\usepackage{graphicx}
\usepackage{float}
\usepackage{caption}
\usepackage{subcaption}
\usepackage{amsmath}
\usepackage{amssymb}
\usepackage{mathtools}
\usepackage{amsthm}
\usepackage{enumitem}
\usepackage{url}
\usepackage[hidelinks]{hyperref}
\usepackage[capitalize,noabbrev]{cleveref}
\usepackage{fontawesome5}   

\newcommand{\hflogo}{\raisebox{-0.22em}{\includegraphics[height=0.95em]{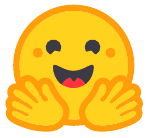}}}

\usepackage{tikz}
\usetikzlibrary{positioning, arrows.meta, backgrounds, calc}
\pgfdeclarelayer{background}
\pgfsetlayers{background,main}

\theoremstyle{plain}

\theoremstyle{definition}

\theoremstyle{remark}

\graphicspath{{figures/}}

\title{FactoryNet: A Large-Scale Dataset toward\\
       Industrial Time-Series Foundation Models}

\author{%
  \textbf{Jonas Petersen}$^{1,2}$\thanks{Corresponding author: \texttt{jep79@cantab.ac.uk}} \quad
  \textbf{Pratik Jawahar}$^{2}$ \quad
  \textbf{Riccardo Maggioni}$^{2}$ \quad
  \textbf{Alexandre Huou}$^{3}$ \\
  \textbf{Karim Othman}$^{2,4}$ \quad
  \textbf{Camilla Mazzoleni}$^{2}$ \quad
  \textbf{Matteo Cozzi}$^{2}$ \quad
  \textbf{Marek Gajewski}$^{2}$ \\
  \textbf{Davide Paloschi}$^{1}$ \quad
  \textbf{Gian-Alessandro Lombardi}$^{2}$ \quad
  \textbf{Matei C. Ignuta-Ciuncanu}$^{5}$ \\
  \textbf{Lior Porath}$^{1}$ \quad
  \textbf{Federico Martelli}$^{1,2}$ \quad
  \textbf{Philipp Christian Petersen}$^{6}$ \\[0.6em]
  $^{1}$ETH Zurich \quad $^{2}$Forgis \quad $^{3}$EPFL \quad $^{4}$Cairo University \quad
  $^{5}$UC Berkeley \quad $^{6}$University of Vienna
}

\iclrfinalcopy

\begin{document}

\maketitle
\lhead{Preprint}

\vspace{-1.25em}
\centerline{\small\faGithub\hspace{0.4em}\href{https://github.com/Forgis-Labs/FactoryNet}{github.com/Forgis-Labs/FactoryNet}\hspace{2em}\hflogo\hspace{0.4em}\href{https://huggingface.co/datasets/Forgis/FactoryNet}{huggingface.co/datasets/Forgis/FactoryNet}}
\vspace{0.7em}

\begin{abstract}
    Industrial machines are actuated systems, yet most open time-series corpora
    record sensor outcomes alone; the few that do log commanded signals expose
    them in vendor-specific forms that differ from machine to machine, so
    \emph{commanded intent} and \emph{measured response} cannot be compared
    across embodiments. We introduce \textbf{FactoryNet}, a pretraining
    corpus for industrial time-series: 113M datapoints across 57k end-to-end
    task executions (15k real, 42k synthetic) on seven embodiments and four
    tasks, with 27 annotated anomaly types alongside healthy baselines and
    counterfactual pairs. Every signal in every source is mapped into a single
    control-theoretic schema, \textit{Setpoint, Effort, Feedback, Context} (S-E-F-C),
    which expresses any actuated system in a common representational frame
    and makes commanded-versus-realized dynamics readable as explicit
    prediction residuals. 
    Using the corpus, we find that a task-phase model trained on simulation, with no real
    phase labels, recovers roughly two thirds of the real-data ceiling on both robots,
    and that transfer holds within a robot family but rarely across manufacturers.
    Fault signatures behave differently: an uncalibrated simulator gives no detectable
    fault signal on either robot, and aligning the feature distributions does not repair
    it. What separation there is comes from \textit{Effort}: motor current distinguishes
    faults where measured tool motion is close to chance, and calibrating the simulator
    to reproduce that one channel recovers a payload fault on the real robot. We read this
    as simulation capturing the stable structure of a task while fault behaviour stays
    diverse and fault-type specific. We release FactoryNet as a growing, multi-embodiment substrate for industrial foundation models.
\end{abstract}

\begin{figure}[t]
    \centering
    \includegraphics[width=\linewidth]{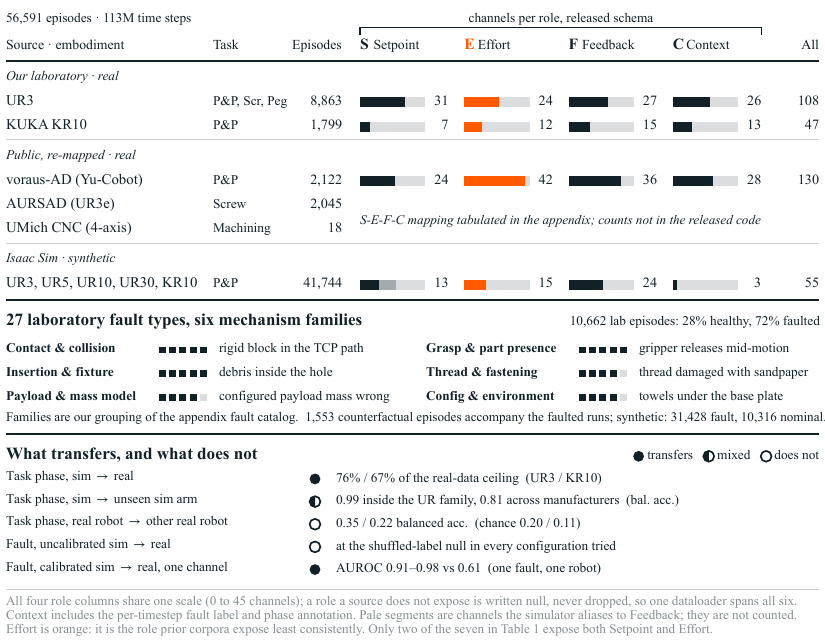}
    \caption{\textbf{FactoryNet at a glance.} 56{,}591 task executions and 113M
    logged time steps across six sources and seven embodiments, every signal
    mapped into one control-theoretic schema. \emph{Top:} channels each source
    exposes per role, on one shared scale; a role a source does not expose is
    written null rather than dropped, so every source loads through the same
    dataloader. Effort, highlighted, is the role prior corpora expose least
    consistently: only two of the seven in
    Table~\ref{tab:dataset_comparison} expose both Setpoint and Effort.
    \emph{Middle:} the
    27 laboratory fault types, grouped by mechanism. \emph{Bottom:} what the
    transfer study (\cref{sec:transfer}) finds carries between domains and what
    does not.}
    \label{fig:dataset_overview}
\end{figure}

\section{Introduction}
The manufacturing sector accounts for approximately 15\% of global GDP,
relying heavily on the continuous operation of complex actuated
machinery~\cite{WB_NV.IND.MANF.ZS}. While predictive maintenance and process
optimization present significant opportunities for machine learning,
industrial AI remains largely confined to single-machine, bespoke
deployments.  Foundation models have transformed vision and language by
pretraining on large, structurally coherent corpora ~\cite{brown2020language, dosovitskiy2020image}, yet no analogous
substrate exists for industrial time-series. The gap is not merely
volume: existing anomaly detection and forecasting
datasets~\cite{Brockmann2023TheVD,leporowski2021aursad} record sensor
outcomes without separating \emph{commanded intent} from
\emph{measured response}.  For actuated systems, learning transferable dynamics requires observing the full control loop from target trajectory through actuation effort to the resulting physical state. Partial solutions exist for specific machines (Section~\ref{sec:related}), but no unified open dataset provides this explicit decomposition across multiple embodiments.\footnote{An \emph{embodiment} is a distinct machine platform, UR3 and UR3e counted as one: UR3/UR3e, UR5, UR10, UR30, KUKA KR10, the voraus-AD cobot, the UMich CNC machining centre.}

Every signal is mapped into the Setpoint-Effort-Feedback-Context
(S-E-F-C) schema: a machine-agnostic signal taxonomy we propose, inspired by the
function-oriented reference designation principles of IEC~81346, so that signals from
different vendors and machine types fall into the same four roles.
Setpoint and Context fix commanded intent and boundary conditions, while Effort
and Feedback make \emph{commanded versus realized} dynamics comparable as
explicit prediction residuals rather than opaque reconstruction scores. Paired
real and simulated episodes under the same schema let sim-to-real mismatch be
read as forward-model error under matched inputs.

The schema makes a split visible. Simulation transfers the \emph{task structure}
of a cycle, meaning which stage is running and in what order, but not its \emph{fault
physics}. S-E-F-C says why: the fault signal sits in \emph{Effort}, the actuation
channel, and Effort is precisely what a kinematic simulator reproduces worst and
what open corpora expose least consistently. Simulation buys commanded motion,
not contact physics. That is a claim about sim-to-real transfer in actuated
systems rather than about robot arms, and it is measurable only because one
schema puts different machines in a common frame.

\section{Related Work}
\label{sec:related}

\subsection{Industrial Fault-Detection Datasets}
The availability of open-source data in the manufacturing domain lags
significantly behind other modalities.  Most existing datasets focus on
single-machine, run-to-failure scenarios or specific rotating machinery
components.  Canonical examples include the NASA C-MAPSS turbofan
degradation benchmark~\cite{saxena2008damage} and recent robotic
datasets: voraus-AD~\cite{Brockmann2023TheVD} provides 2,122 episodes of
industrial robot recordings and AURSAD~\cite{leporowski2021aursad} offers
2,045 episodes.  However, these datasets are bounded in scale and lack an
explicit control-loop structure mapping commands to outcomes.  FactoryNet unifies these sources with novel laboratory data under one control-theoretic decomposition (Table~\ref{tab:dataset_comparison}).

Beyond single-component datasets, broader industrial anomaly detection benchmarks such as SKAB~\cite{skab}, MetroPT~\cite{veloso2022benchmarkdatasetpredictivemaintenance}, WADI~\cite{WADI2017testbed} and the Tennessee-Eastman process~\cite{downs1993plant} provide system-level multivariate time-series, and general-purpose benchmarks such as TSB-AD~\cite{liu2024elephant}, Anomaly-Transformer~\cite{xu2022anomaly} and TODS~\cite{lai2025todsautomatedtimeseries} evaluate or automate detection methods. However, these datasets record holistic system states without the explicit commanded-versus-measured structural decomposition necessary for learning robotic dynamics.

\begin{table}[t]
    \centering
    \caption{Comparison of open-source industrial time-series datasets. FactoryNet is, to our knowledge, the only multi-machine corpus that requires a strict control-loop structure decoupling intended Setpoint from applied Effort. The Eps.\ column is not a like-for-like count: $^{r}$~continuous run-to-failure or steady-state recordings, counted as files, for which any episode boundary is a segmentation choice; $^{c}$~machining cuts (six cutters $\times$ 315 cuts); $^{m}$~measurement runs (32 bearings $\times$ 4 operating conditions $\times$ 20 repeats); all others, and FactoryNet, are end-to-end task executions. The Setpoint and Effort columns, not the count, are what distinguishes the corpora.}
    \label{tab:dataset_comparison}
    \footnotesize
    \setlength{\tabcolsep}{2.5pt}%
        \begin{tabular}{lcllccc}
            \toprule
            \textbf{Dataset}                    & \textbf{Year} & \textbf{Machine Type}  & \textbf{Signals}    & \textbf{Eps.} & \textbf{Setpoint}  & \textbf{Effort}    \\
            \midrule
            CWRU~\cite{cwru}                    & 2000          & Bearings               & Vibration           & 161$^{r}$               & No                      & No                      \\
            PHM 2010~\cite{phm2010}             & 2010          & CNC mill            & Force, vib.    & 1,890$^{c}$             & Partial                 & Yes                     \\
            Paderborn~\cite{paderborn}          & 2016          & Bearings               & Vib., current  & 2,560$^{m}$             & RPM only           & Partial                 \\
            MAFAULDA~\cite{mafaulda}            & 2016          & Rotating mach.     & Vib., audio    & 1,951$^{r}$             & No                      & No                      \\
            XJTU-SY~\cite{xjtu}                 & 2019          & Bearings               & Vibration           & 15$^{r}$                & No                      & No                      \\
            AURSAD~\cite{leporowski2021aursad}  & 2021          & UR3e robot             & Joint sig.       & 2,045                   & Yes                     & Yes                     \\
            voraus-AD~\cite{Brockmann2023TheVD} & 2024          & Cobot    & 130 channels        & 2,122                   & Yes                     & Yes                     \\
            \midrule
            \textbf{FactoryNet (Ours)}          & \textbf{2026} & \textbf{Multi-machine} & \textbf{Multimodal} & \textbf{56,591}            & \textbf{Required} & \textbf{Required} \\
            \bottomrule
        \end{tabular}
\end{table}

\subsection{Foundation Models for Time-Series and Robotics}
Current foundation model research diverges into two distinct tracks.  In
robotics, Open X-Embodiment~\cite{o2024open} pools data across diverse
kinematics, and DROID~\cite{khazatsky2024droid} across diverse scenes and
tasks on a single platform, to train generalizable behavioral policies.
In the structured time-series domain, models such as
Chronos~\cite{ansari2024chronos}, TimesFM~\cite{das2023decoder},
Moirai~\cite{woo2024unified}, Timer~\cite{liu2024timer} and
Lag-Llama~\cite{rasul2023lag} leverage pretraining corpora for zero-shot
forecasting, while MOMENT~\cite{pmlr-v235-goswami24a} targets broader analysis and
adaptation and TabPFN-TS~\cite{hoo2025tables} adapts a tabular prior-fitted network
with no time-series pretraining. However, because these
TS-FMs are predominantly trained on web-scraped, financial, or ecological data, we
hypothesize that they lack grounding in physical actuation and may struggle
to disentangle static payload biases from dynamic behavior.  FactoryNet bridges
these tracks: an industrial-scale, control-theoretically structured corpus for training and evaluating TS-FMs on complex dynamics across embodiments.

\subsection{Physics-Informed and Structured Dynamics}
In parallel to data-driven forecasting, Physics-Informed Neural Networks (PINNs)~\cite{RAISSI2019686} and grey-box models incorporate known governing equations into the learning process. While these approaches excel in scenarios with well-defined partial differential equations, complex multi-joint robotic systems are commonly approached with structured priors or residual-dynamics learning, as in TossingBot~\cite{zeng2020tossingbot}. FactoryNet complements this literature with the empirical Setpoint-Effort-Feedback decomposition needed to train and evaluate such models at scale.

\section{The FactoryNet Dataset}
FactoryNet v1.0 is to our knowledge, the largest open-source industrial-robot time-series
dataset containing labelled anomalies and organized around a
control-theoretic schema.  It comprises three pillars: real-world
laboratory recordings, standardized open-source adaptations, and a
synthetic generation pipeline.

\subsection{Dataset Composition}
\begin{figure}[t]
    \centering
    \includegraphics[width=0.92\linewidth]{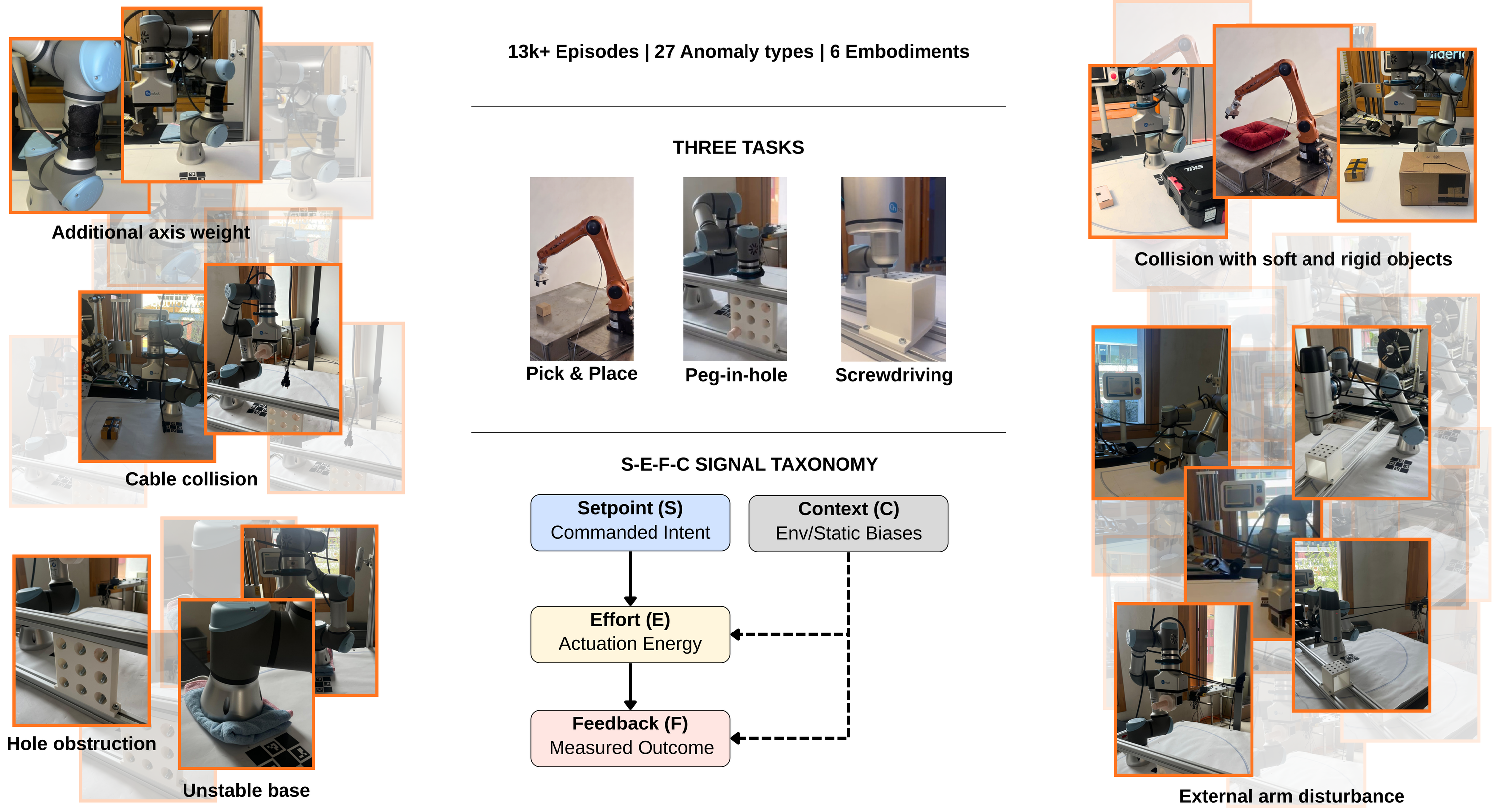}
    \caption{\textbf{FactoryNet overview.} A multi-embodiment industrial time-series
    corpus spanning 57k episodes (14.8k real, 41.7k simulated) on seven embodiments,
    27 anomaly types, and three manipulation tasks. Every signal is mapped into the
    Setpoint-Effort-Feedback-Context (S-E-F-C) taxonomy, a control-theoretic
    decomposition that enables cross-embodiment learning. Representative fault types
    are shown alongside the corresponding laboratory setups.}
    \label{fig:visual_abstract}
\end{figure}

As detailed in Table \ref{tab:dataset_composition} and Figure~\ref{fig:dataset_overview}, the corpus encompasses multiple data streams. The laboratory track consists of novel recordings we collected from UR3 and KUKA KR10
industrial robotic arms executing three complex tasks: Pick~\&~Place, Screwdriving,
and Peg-in-Hole; \cref{fig:visual_abstract} shows the cells and representative
injected faults. We ingested and restructured existing high-quality
open-source datasets (voraus-AD, AURSAD, and UMich CNC~\cite{sharon_2017_toolwear}) into our unified
schema, which also accommodates CNC milling centres and rotating machinery.  A parallel synthetic track generated via NVIDIA Isaac
Sim provides procedurally scaled data for model pretraining.

\begin{table}[!t]
    \centering
    \caption{FactoryNet composition. Episodes are normal and fault recordings
    (1,553 real counterfactual episodes come in addition). Datapoints are
    logged time steps and are approximate.}
    \label{tab:dataset_composition}
    \resizebox{\columnwidth}{!}{%
        \small
        \begin{tabular}{llccrr}
            \toprule
            \textbf{Source}         & \textbf{Machine}     & \textbf{Tasks}
                                    & \textbf{Faults}      & \textbf{Episodes}
                                    & \textbf{Datapoints}                            \\
            \midrule
            \textbf{Our Lab (Real)} & UR3                  & P\&P, Screw, Peg  & Yes
                                    & 8,863                & 11M                     \\
            \textbf{Our Lab (Real)} & KUKA KR10            & P\&P              & Yes
                                    & 1,799                & 4M                      \\
            \midrule
            \textbf{Open (Real)}    & voraus-AD (Yu-Cobot) & P\&P              & Yes
                                    & 2,122                & 16M                     \\
            \textbf{Open (Real)}    & AURSAD (UR3e)        & Screw             & Yes
                                    & 2,045                & 3M                      \\
            \textbf{Open (Real)}    & UMich CNC            & Machining         & Yes
                                    & 18                   & 18K                     \\
            \midrule
            \textbf{Synthetic}      & Isaac Sim (5 arms)   & P\&P              & Yes
                                    & 41,744               & 79M                     \\
            \midrule
            \textbf{Total}          &                      &                   &
                                    & 56,591               & 113M                    \\
            \bottomrule
        \end{tabular}%
    }
\end{table}


\subsection{The S-E-F-C Signal Taxonomy: A Unified Vocabulary}
The defining feature of FactoryNet is its control-theoretic structure.
Most existing benchmarks simply log raw sensor streams, intrinsically tangling the controller's target variables with the machine's actual physical execution. This conflates cause and effect while suffering from vendor-specific naming conventions.

To resolve this, we employ embodiment-specific adapter scripts to programmatically map over 300 heterogeneous data columns into four standardized roles (machine-readable tables: Appendix~\ref{app:sefc_mapping}). These categories are: \textbf{Setpoint (S)} (commanded intent, e.g., target joint positions), \textbf{Effort (E)} (actuation energy expended, e.g., motor current or torque, where exposed), \textbf{Feedback (F)} (measured physical outcome, e.g., actual positions), and \textbf{Context (C)} (environmental or static variables, e.g., payload mass).

This taxonomy allows a single dataloader to work across UR3, KUKA KR10,
and CNC machinery: a 6-DOF rotational arm and a 4-axis CNC gantry
expose the same four roles, only with different axis counts and units.
By enforcing this taxonomy, S-E-F-C acts as a unified vocabulary for
cross-embodiment models, providing a principled inductive bias for
learning the difference between expected dynamics and external
disturbances.  Signal availability varies by embodiment: KUKA KR10
(KSS~8.3) does not expose joint velocities, commanded TCP pose, or TCP
force/torque via its RSI interface; these channels are marked absent in
the S-E-F-C mapping (Appendix~\ref{app:sefc_mapping}).

\subsection{Synthetic Pipeline and Sim-to-Real}
To reduce dependence on real-only data, FactoryNet includes a synthetic
pipeline in NVIDIA Isaac Sim 5.1 / Isaac Lab 2.3 (GPU-batched, PhysX at
500~Hz). It yields 41,744 validated Pick~\&~Place episodes across five arms
(UR3, UR5, UR10, UR30, KUKA KR10) with aligned S-E-F-C metadata: 10,316
nominal and 31,428 fault episodes spanning six fault classes. Control and
logging run at 125~Hz (UR) and 83.3~Hz (KR10, the KRC interpolation cycle),
with no resampling. The UR3 and KR10 motion profiles (velocity,
acceleration, path types) are fitted to real recordings; the UR5, UR10 and
UR30 profiles have no real counterpart and are scaled from the UR3 fit, an
assumption we make explicit.
\begin{itemize}
    \item \textbf{Payload and contact:} part mass is sampled per episode from
    $U(0.30, 1.20)$~kg (UR) and $U(0.30, 2.50)$~kg (KR10). Part friction is
    sampled from $U(0.6, 1.2)$ against a fixed pad coefficient of $1.0$.
    Robot link masses and arm drive gains are fixed. The gripper gain is
    derived per episode from the required grip force rather than sampled.
    \item \textbf{Task and geometry:} pick and place poses are sampled within
    each station's workspace with random yaw, and part dimensions are
    randomized within bounds. Execution speed is scaled by $U(0.76, 1.24)$.
    \item \textbf{Faults:} six classes (payload addition, motor
    miscommutation, gripper activation failure, gripper release, collision,
    path obstacle) are injected at a phase drawn uniformly among the valid
    phases of each episode (Appendix~\ref{app:sim_validity}).
    \item \textbf{Physics and control:} each arm is an Isaac Lab articulation
    with a welded Robotiq 2F-85 gripper, driven by joint-space PD controllers
    (critically damped; stiffness fixed per arm and scaled from the UR3 values
    by torque limit). A software model of the vendor motion layer sits above
    them: trapezoidal joint and linear profiles, analytic inverse kinematics,
    computed-torque feedforward (gravity, Coriolis, inertia) and a joint-friction
    prior. The UR3 and KR10 profiles are fitted to 109 and 280 real nominal
    episodes. This is a simplified stand-in for the proprietary UR and KRC
    controllers, not a copy of their control law, and it has known deviations
    (for example, the logged tracking error is inflated relative to the real
    robot); gains, limits and deviations are listed in
    Appendix~\ref{app:sim_physics}.
\end{itemize}

\paragraph{Validity gate.}
Each generated episode passes a two-stage gate before release. The
structural stage requires a plausible length, complete task phases, a
placed part for nominal episodes, the injected event present in the log, and
non-zero contact force. The trace stage requires every fault episode to
leave a measurable physical trace, such as an outcome failure, a torque
excursion above the nominal envelope, a stalled grip, or a protective stop;
otherwise it is quarantined. Of 50,000 generated episodes, 41,744 pass
(nominal 10,316 of 12,500; fault 31,428 of 37,500). Most fault rejections
(5,138 of 6,072) are faults that left no trace. This choice ensures that
every released fault label corresponds to a physical effect, but it also
biases the synthetic faults toward more detectable instances, and acceptance
rates differ across fault classes (e.g., motor miscommutation passes far less
often than gripper faults). Motor miscommutation is excluded for the KR10,
where its effect is sub-threshold. Per-arm acceptance rates are given in
Appendix~\ref{app:sim_validity}.

\paragraph{Measured sim-to-real gap.}
An earlier UR3 campaign paired 1{,}155 real and simulated episodes and measured
the per-phase gap directly (Appendix~\ref{app:sim2real_gap}); it used Isaac Sim
4.5.0 and a different gripper from the laboratory setup, so its absolute pose
errors are conservative. Whether the simulator is close enough for a \emph{model}
to transfer is the question Section~\ref{sec:transfer} tests directly, and is the
one that matters for this corpus.

\subsection{Faults and Anomalies}
To support anomaly detection and robust control research, of the
10,662 laboratory normal and fault episodes, approximately 28\% are healthy and 72\% contain injected faults across 27 anomaly types spanning three tasks: \textbf{Pick \& Place}, \textbf{Screwdriving}, and \textbf{Peg-in-Hole}. A further 1,553 real counterfactual episodes accompany the faulty runs.

\subsection{Data Accessibility and Licensing}
Novel laboratory and synthetic data are released under the MIT license;
adapted subsets retain their original licenses, listed per source in
Appendix~\ref{app:datasheet}, with voraus-AD under CC~BY-NC-SA~4.0, so that subset
stays NonCommercial and ShareAlike. The Hugging Face dataset linked on the first page provides
S-E-F-C Parquet files, metadata and framework-native dataloaders: at submission it
serves every real subset in final form and the simulation data behind the reported
experiments, while the full Isaac Lab synthetic campaign is still uploading during
review. The verification script reproduces every reported value from stored result
files, without the corpus download.

\section{Dataset Utility \& Validation}
We ask two questions of the corpus. First, does the S-E-F-C schema support
competitive anomaly detection on a single machine with far fewer signals?
Second, which parts of a monitoring model carry over from simulation to real
robots and between real robots? The first is a baseline that shows the schema is
usable; the second is the main study. Confidence intervals are 95\% bootstrap
intervals over episodes unless stated otherwise.

\subsection{Single-Machine Baseline: Validating the S-E-F-C Schema}
\label{sec:exp_details}
We use voraus-AD to test whether S-E-F-C enables competitive anomaly detection via supervised dynamics rather than holistic reconstruction. We follow the official protocol of \citet{Brockmann2023TheVD}: training on 948 healthy episodes only, testing on the 1,174-episode labelled set (419 healthy, 755 anomalous). Unstructured baselines reconstruct all 130 channels end-to-end and use the architectures and hyperparameters of the original benchmark. Our \textbf{S-E-F-C MLP} is a regressor on 24 signals: it maps 18 Setpoints (\texttt{setpoint\_pos\_0\dots5}, \texttt{setpoint\_vel\_0\dots5}, \texttt{setpoint\_acc\_0\dots5}) to 6 Efforts (\texttt{effort\_motor\_torque\_0\dots5}), and the per-episode mean absolute error on Effort is the anomaly score. The network has three hidden layers (512, 256, 128; ReLU, no dropout) and is trained with Adam (learning rate $5\times10^{-4}$, weight decay $10^{-5}$, batch size 4,096, cosine annealing, up to 500 epochs, early stopping with patience 30). On 24 signals alone it reaches \textbf{83.2\%} mean AUROC. Table~\ref{tab:anomaly_baselines} shows it beats weaker full-channel baselines (1-NN, GANF, PCA) but not the strongest ones (CAE, LSTM-VAE, HMM, MVT-Flow). Per-category results are in Appendix~\ref{app:per_category}: the MLP is strongest on mechanically distinctive faults (miscommutation 99.2, additional axis weight 95.8) and weakest on brief gripping failures (collision with cables 67.6, losing can 71.8), where a single-step model has no temporal context.

\begin{table}[H]
    \centering
    \caption{Mean AUROC on voraus-AD. Rows marked (Ours) are this work's own results; all other values are quoted from~\citet{Brockmann2023TheVD}. Baselines use 130 channels, our S-E-F-C MLP 24 (Setpoint$\to$Effort). $^{\ddagger}$Std across 12 fault categories.}
    \label{tab:anomaly_baselines}
    \small
    \setlength{\tabcolsep}{6pt}%
    \begin{tabular}{lcc}
        \toprule
        \textbf{Method}             & \textbf{Input Channels} & \textbf{Mean AUROC}                  \\
        \midrule
        MLP - All signals (Ours)    & 130                     & 69.1 $\pm$ 3.1$^{\ddagger}$          \\
        1-NN                        & 130                     & 77.5                                 \\
        GANF                        & 130                     & 79.9 $\pm$ 12.7$^{\ddagger}$         \\
        PCA                         & 130                     & 80.0                                 \\
        \textbf{S-E-F-C MLP (Ours)} & \textbf{24}             & \textbf{83.2 $\pm$ 9.6$^{\ddagger}$} \\
        CAE                         & 130                     & 85.2 $\pm$ 9.2$^{\ddagger}$          \\
        LSTM-VAE                    & 130                     & 86.7 $\pm$ 10.1$^{\ddagger}$         \\
        HMM                         & 130                     & 87.4 $\pm$ 5.8$^{\ddagger}$          \\
        MVT-Flow                    & 130                     & 93.6 $\pm$ 5.7$^{\ddagger}$          \\
        \bottomrule
    \end{tabular}
    \normalsize
\end{table}

\subsection{Which S-E-F-C Roles Carry the Signal}
\label{sec:sefc_ablation}
The schema is our central claim, so we test it directly rather than by a single
number. Holding the architecture, the training set and the evaluation protocol of
Section~\ref{sec:exp_details} fixed, we vary only which roles supply the inputs and
which supply the prediction target, and run five seeds per variant
(Table~\ref{tab:sefc_ablation}).

\begin{table}[H]
    \centering
    \caption{S-E-F-C role ablation on voraus-AD, official split, identical
    architecture and protocol throughout; only the role assignment changes. Mean
    AUROC over the 12 fault categories, $\pm$ standard deviation across 5 seeds
    (10 draws for the random baseline). MVT-Flow, the strongest published
    baseline, scores 93.6 on 130 channels.}
    \label{tab:sefc_ablation}
    \small
    \begin{tabular}{llcc}
        \toprule
        \textbf{Inputs} & \textbf{Target} & \textbf{Channels} & \textbf{Mean AUROC} \\
        \midrule
        Setpoint (pos, vel, acc)      & Feedback (torque sensors) & 18$\to$6  & \textbf{94.8 $\pm$ 0.3} \\
        Feedback + Context            & Effort (motor torque)     & 64$\to$6  & 89.5 $\pm$ 0.7 \\
        Setpoint + Feedback           & Effort                    & 60$\to$6  & 87.9 $\pm$ 0.6 \\
        All non-Effort roles          & Effort                    & 88$\to$6  & 86.4 $\pm$ 1.0 \\
        \textit{Random 18 of the other 124} & Effort              & 18$\to$6  & \textit{83.3 $\pm$ 5.2} \\
        Setpoint (pos, vel, acc)      & Effort \emph{(Sec.~\ref{sec:exp_details})} & 18$\to$6 & 82.8 $\pm$ 0.6 \\
        All 124 non-target signals    & Effort                    & 124$\to$6 & 76.2 $\pm$ 0.8 \\
        Setpoint (pos, vel, acc)      & Feedback (joint pos, vel) & 18$\to$12 & 61.3 $\pm$ 0.2 \\
        Context only                  & Effort                    & 28$\to$6  & 60.9 $\pm$ 1.9 \\
        \bottomrule
    \end{tabular}
\end{table}

\paragraph{Role choice matters more than channel count.}
Across the role assignments and channel budgets we evaluate, mean AUROC varies by
33.9 points, and the variant that sees \emph{every} available signal is the third
worst: adding channels without regard to their role actively hurts. The target role
dominates: from the same 18 commanded-motion inputs, predicting measured torque
reaches 94.8, above the strongest published voraus-AD baseline, predicting motor
Effort 82.8 (a 12.0-point drop at an identical 18$\to$6 budget), and predicting
measured \emph{kinematics} 61.3, because a position residual carries almost no
information about load while a torque residual carries most of it.

\paragraph{The taxonomy is a search space, not a guarantee.}
The particular pairing used in Section~\ref{sec:exp_details} is not itself special:
a random 18-channel input set scores the same on average (83.3 against 82.8). The
two deviations are not comparable (5.2 across channel draws, 0.6 across seeds at one
fixed assignment), and the substantive point is the ceiling: no random draw reaches
the Setpoint$\to$Feedback result. We therefore
read S-E-F-C as a small structured space of role pairings that is cheap to search
and contains a state-of-the-art point, not as a claim that any one pairing is good.
This also bounds the claim: the ablation covers one benchmark and one architecture
family, and the ranking of roles on other machines is untested.

\subsection{Transferability Study: Simulation to Real and Robot to Robot}
\label{sec:transfer}
We use FactoryNet to ask what a monitoring model learns in simulation, or on
another robot, that still works on a real machine. We test two tasks: phase
identification (which of the ten stages of the pick-and-place cycle is running)
and binary fault detection. Both use MiniRocket, a fixed convolutional-kernel transform from the ROCKET
family that is not pretrained \citep{dempster2021minirocket}, with a ridge classifier,
and are compared against shuffled-label nulls and bootstrap intervals. Real test
sets are single sessions, for the reason given under Caveats below. Protocol and
all cells are in Appendix~\ref{app:transfer}.

\begin{figure}[!t]
    \centering
    \includegraphics[width=\linewidth]{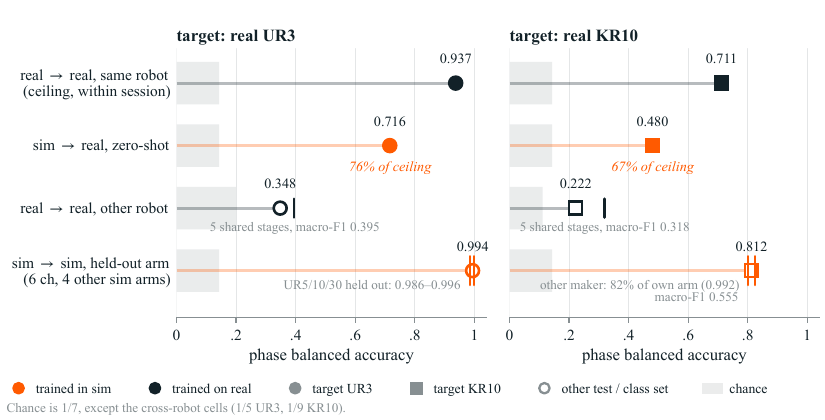}
    \caption{\textbf{Task phase transfers from simulation; between real robots it
        does not.} 7-stage balanced accuracy on the real UR3 and KR10: the
        within-session real ceiling, sim$\to$real zero-shot, real cross-robot
        transfer (tick: macro-F1 on the five shared stages) and sim$\to$sim
        leave-one-arm-out on the same six channels. Whisker = 95\% bootstrap
        interval. A model trained with no real phase labels recovers
        76\% (UR3) and 67\% (KR10) of the real-data ceiling on the sessions that emit
        all seven stages -- 68\% and 66\% pooling every session
        (\cref{app:transfer_phase}) -- while a model
        trained on the other \emph{real} robot does markedly worse.}
    \label{fig:transfer_phase}
\end{figure}

\begin{figure}[!t]
    \centering
    \includegraphics[width=\linewidth]{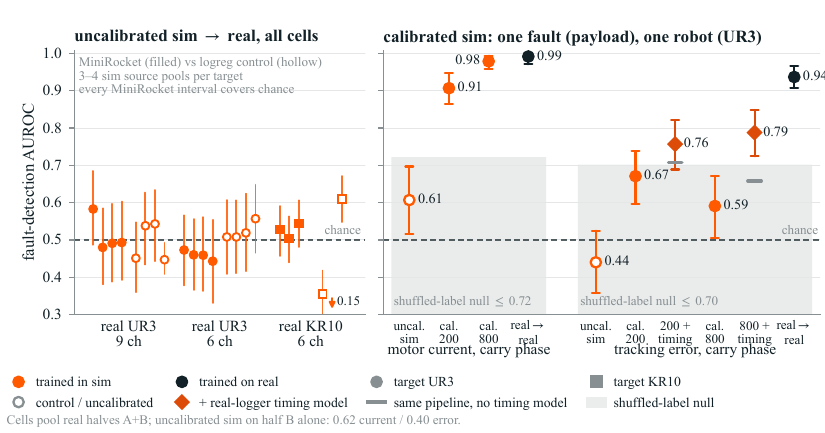}
    \caption{\textbf{Fault signatures do not transfer from uncalibrated
        simulation, and calibrating one channel recovers one fault.} Fault
        AUROC on real data. Left: uncalibrated sim$\to$real, in every
        configuration we tried. Right: simulation calibrated to a single
        payload fault on a single robot (UR3), with and without the real
        logger's timing model, against the real$\to$real ceiling. ``cal.\ 200''
        and ``cal.\ 800'' are calibration campaigns of 200 and 800 simulated
        episodes. The right-hand panel is a case study, not a general result.}
    \label{fig:transfer_fault}
\end{figure}

\paragraph{Task phase carries over from simulation.}
A model trained on simulation, with no real phase labels, recovers roughly three
quarters of the accuracy of a model trained on real data (76\% on the UR3, 67\% on
the KR10; \cref{fig:transfer_phase}). Its confusions are systematic rather than
random, which suggests that it captures the order of the stages better than the
timing of real recordings. Between simulated arms, transfer to an unseen arm
is strong inside the UR family and weaker across manufacturers.

\paragraph{Between real robots, transfer is limited.}
A phase model trained on one real robot transfers only weakly to the other
(\cref{fig:transfer_phase}), while the same robot transfers reasonably across
programs and sessions. The stage labels appear to mark the same physical events on
both robots, whereas the task programs differ in cycle time, pick-to-place
distance and per-stage speed. Simulation does not separate the explanations:
with a shared task program, simulated UR3 and KR10 transfer to each other only
weakly as well. We therefore hypothesise that the gap reflects task program and
embodiment jointly; our data do not tell them apart.

\paragraph{Fault signatures appear to be more specific.}
Uncalibrated simulation gives no detectable fault signal on the real robots in
any configuration we tried, and aligning the feature distributions does not
change this (\cref{fig:transfer_fault}). Real-to-real detection works within a
session for fault types seen in training, and is near chance for unseen types.
This suggests, tentatively, that the models pick up fault-specific signatures
rather than a generic anomaly direction; adding real episodes from other sessions
did not raise the ceiling.

\paragraph{Calibration may help where the effort channel is matched.}
As a first probe we calibrated the simulator to one fault, a payload change
on the UR3. A detector trained on simulation alone detects it in real motor
current (AUROC 0.91--0.98, against 0.61 uncalibrated), and in joint tracking
error only once the simulator also models the real logger's timing
(0.76--0.79, against 0.59--0.67). One reading is that sim-trained detectors
transfer when the simulator reproduces the physics of the channel they rely
on. It is one fault on one robot, calibrated with real episodes of that fault and
not carrying over to the KR10: a case study.

\paragraph{Caveats.}
Recording program, day and speed override partly predict the fault label on both
robots, so we test within single sessions; only two of the four KR10 fault types
are free of a speed confound, and the clean test sets are small. In the
calibrated case the real classes also differ in gripper dwell, so a class-linked
recording effect cannot be fully excluded. UR5, UR10 and UR30 exist only in
simulation, so their evidence is simulation to simulation.


\vspace{-0.5em}
\section{Conclusion \& Outlook}
\vspace{-0.5em}
FactoryNet provides a large open corpus of fault-injected industrial-robot time
series, real and simulated, unified by S-E-F-C. The schema is not
cosmetic: at a fixed architecture and a matched 18$\to$6 channel budget, which roles
predict which is worth 12 AUROC points on voraus-AD. Our transfer study suggests that task phase
may be a stable structure simulation can capture, recovering roughly two thirds
of the real-data ceiling with no real phase labels, that this holds inside a
robot family more than across manufacturers, and that fault signatures are more
specific: uncalibrated simulation gave no detectable fault signal on real robots,
while a simulator calibrated to one payload fault detected it on the real UR3.
These are hypotheses, not settled explanations.

Each boundary marks work the corpus makes possible. Recording the UR5, UR10 and
UR30, today simulation-only, would make the multi-arm evidence real; a task
program matched across manufacturers would separate program, tempo and
embodiment as the cause of weak cross-robot transfer; session-balanced
collection would remove the recording context that predicts fault labels. None of
that needs a new dataset, only the same schema on more machines that share no vendor,
kinematics or sensor set: the step from models of one machine toward foundation
models for industry.

\clearpage

\bibliography{factorynet}

@article{Brockmann2023TheVD,
  title={The {voraus-AD} Dataset for Anomaly Detection in Robot Applications},
  author={Jan Thie{\ss} Brockmann and Marco Rudolph and Bodo Rosenhahn and Bastian Wandt},
  journal={{IEEE} Transactions on Robotics},
  year={2024},
  volume={40},
  pages={438--451},
  doi={10.1109/TRO.2023.3332224},
  url={https://ieeexplore.ieee.org/document/10315239}
}

@article{leporowski2021aursad,
  title={{AURSAD}: Universal Robot Screwdriving Anomaly Detection Dataset},
  author={Leporowski, B{\l}a{\.z}ej and Tola, Daniella and Hansen, Casper and Iosifidis, Alexandros},
  journal={arXiv},
  year={2021},
  eprint={2102.01409},
  archivePrefix={arXiv},
  primaryClass={cs.RO},
  url={https://arxiv.org/abs/2102.01409}
}

@inproceedings{o2024open,
  title={Open {X-Embodiment}: Robotic Learning Datasets and {RT-X} Models},
  author={O'Neill, Abby and Rehman, Abdul and Maddukuri, Abhiram and Gupta, Abhishek and Padalkar, Abhishek and Lee, Abraham and Pooley, Acorn and Gupta, Agrim and Mandlekar, Ajay and Jain, Ajinkya and others},
  booktitle={2024 {IEEE} International Conference on Robotics and Automation (ICRA)},
  pages={6892--6903},
  year={2024},
  organization={{IEEE}},
  doi={10.1109/ICRA57147.2024.10611477},
  url={https://ieeexplore.ieee.org/document/10611477}
}

@inproceedings{khazatsky2024droid,
  author    = {Alexander Khazatsky and Karl Pertsch and Suraj Nair and others},
  title     = {DROID: A Large-Scale In-The-Wild Robot Manipulation Dataset},
  booktitle = {Proceedings of Robotics: Science and Systems},
  year      = {2024},
  address   = {Delft, Netherlands},
  month     = {July},
  url       = {https://www.roboticsproceedings.org/rss20/p120.html}
}

@article{ansari2024chronos,
  title={Chronos: Learning the language of time series},
  author={Ansari, Abdul Fatir and Stella, Lorenzo and Turkmen, Caner and Zhang, Xiyuan and Mercado, Pedro and Shen, Huibin and Shchur, Oleksandr and Rangapuram, Syama Sundar and Pineda Arango, Sebastian and Kapoor, Shubham and others},
  journal={Transactions on Machine Learning Research},
  year={2024},
  url={https://openreview.net/forum?id=gerNCVqqtR}
}

@inproceedings{das2023decoder,
  title={A decoder-only foundation model for time-series forecasting},
  author={Das, Abhimanyu and Kong, Weihao and Sen, Rajat and Zhou, Yichen},
  booktitle={Proceedings of the 41st International Conference on Machine Learning},
  series={Proceedings of Machine Learning Research},
  volume={235},
  pages={10148--10167},
  publisher={PMLR},
  year={2024},
  url={https://proceedings.mlr.press/v235/das24c.html}
}

@inproceedings{woo2024unified,
  title={Unified training of universal time series forecasting transformers},
  author={Woo, Gerald and Liu, Chenghao and Kumar, Akshat and Xiong, Caiming and Savarese, Silvio and Sahoo, Doyen},
  booktitle={Forty-first International Conference on Machine Learning},
  series={Proceedings of Machine Learning Research},
  volume={235},
  pages={53140--53164},
  publisher={PMLR},
  year={2024},
  url={https://proceedings.mlr.press/v235/woo24a.html}
}

@article{hoo2025tables,
  title={From Tables to Time: Extending {TabPFN-v2} to Time Series Forecasting},
  author={Hoo, Shi Bin and M{\"u}ller, Samuel and Salinas, David and Hutter, Frank},
  journal={arXiv preprint arXiv:2501.02945},
  year={2025}
}

@inproceedings{saxena2008damage,
  title={Damage propagation modeling for aircraft engine run-to-failure simulation},
  author={Saxena, Abhinav and Goebel, Kai and Simon, Don and Eklund, Neil},
  booktitle={2008 International Conference on Prognostics and Health Management},
  pages={1--9},
  year={2008},
  organization={{IEEE}},
  doi={10.1109/PHM.2008.4711414},
  url={https://ieeexplore.ieee.org/document/4711414}
}

@InProceedings{pmlr-v235-goswami24a,
  title = 	 {{MOMENT}: A Family of Open Time-series Foundation Models},
  author =       {Goswami, Mononito and Szafer, Konrad and Choudhry, Arjun and Cai, Yifu and Li, Shuo and Dubrawski, Artur},
  booktitle = 	 {Proceedings of the 41st International Conference on Machine Learning},
  pages = 	 {16115--16152},
  year = 	 {2024},
  editor = 	 {Salakhutdinov, Ruslan and Kolter, Zico and Heller, Katherine and Weller, Adrian and Oliver, Nuria and Scarlett, Jonathan and Berkenkamp, Felix},
  volume = 	 {235},
  series = 	 {Proceedings of Machine Learning Research},
  month = 	 {21--27 Jul},
  publisher =    {PMLR},
  url = 	 {https://proceedings.mlr.press/v235/goswami24a.html}
}

@inproceedings{liu2024timer,
title={Timer: Generative Pre-trained Transformers Are Large Time Series Models},
author={Liu, Yong and Zhang, Haoran and Li, Chenyu and Huang, Xiangdong and Wang, Jianmin and Long, Mingsheng},
booktitle={Proceedings of the 41st International Conference on Machine Learning},
series={Proceedings of Machine Learning Research},
volume={235},
pages={32369--32399},
publisher={PMLR},
year={2024},
url={https://proceedings.mlr.press/v235/liu24cb.html}
}

@article{rasul2023lag,
  title={Lag-Llama: Towards Foundation Models for Probabilistic Time Series Forecasting},
  author={Rasul, Kashif and Ashok, Arjun and Williams, Andrew Robert and Ghonia, Hena and Bhagwatkar, Rishika and Khorasani, Arian and Bayazi, Mohammad Javad Darvishi and Adamopoulos, George and Riachi, Roland and Hassen, Nadhir and others},
  journal={arXiv},
  year={2023},
  eprint={2310.08278},
  archivePrefix={arXiv},
  primaryClass={cs.LG},
  url={https://arxiv.org/abs/2310.08278}
}

@misc{cwru,
  author       = {Loparo, Kenneth A.},
  title        = {Case {Western} {Reserve} {University} {B}earing {D}ata {C}enter},
  year         = {2000},
  howpublished = {\url{https://engineering.case.edu/bearingdatacenter}},
  note         = {Accessed 2026-09-26. The data centre states no collection or
                  publication date; 2000 follows common usage in this literature}
}

@misc{phm2010,
  author       = {{PHM Society}},
  title        = {2010 {PHM} {S}ociety {C}onference {D}ata {C}hallenge},
  year         = {2010},
  howpublished = {\url{https://phmsociety.org/phm_competition/2010-phm-society-conference-data-challenge/}},
  note         = {Dataset and challenge page for remaining useful life estimation of high-speed CNC milling machine cutters; accessed 2026-05-08}
}

@inproceedings{paderborn,
  title={Condition monitoring of bearing damage in electromechanical drive systems by using motor current signals of electric motors: a benchmark data set for data-driven classification},
  author={Lessmeier, Christian and Kimotho, James Kuria and Zimmer, Detmar and Sextro, Walter},
  booktitle={Proceedings of the European Conference of the Prognostics and Health Management Society},
  year={2016}
}

@misc{mafaulda,
  title={{MAFAULDA} :: Machinery Fault Database},
  author={Marins, Matheus A. and Ribeiro, Felipe M. L. and Netto, Sergio L. and {da Silva}, Eduardo A. B.},
  year={2016},
  note={Signals, Multimedia, and Telecommunications Laboratory, COPPE/Poli/UFRJ; contact: felipe.ribeiro@smt.ufrj.br},
  howpublished={\url{http://www02.smt.ufrj.br/~offshore/mfs/page_01.html}}
}

@article{xjtu,
  title  = {{XJTU-SY} Rolling Element Bearing Accelerated Life Test Datasets: A Tutorial},
  author = {Lei, Yaguo and Han, Tianyu and Wang, Biao and Li, Naipeng and Yan, Tao and Yang, Jun},
  journal = {Journal of Mechanical Engineering},
  volume  = {55},
  number  = {16},
  pages   = {1--6},
  year    = {2019},
  doi     = {10.3901/JME.2019.16.001}
}

@article{gebru2021datasheets,
  title={Datasheets for datasets},
  author={Gebru, Timnit and Morgenstern, Jamie and Vecchione, Briana and Vaughan, Jennifer Wortman and Wallach, Hanna and Daum{\'e} III, Hal and Crawford, Kate},
  journal={Communications of the ACM},
  volume={64},
  number={12},
  pages={86--92},
  year={2021},
  publisher={ACM},
  doi={10.1145/3458723},
  url={https://doi.org/10.1145/3458723}
}

@misc{WB_NV.IND.MANF.ZS,
  author = {{World Bank}},
  title = {Manufacturing, value added (\% of {GDP}): {World} aggregate},
  year = {2026},
  organization = {World Bank Open Data},
  url = {https://data.worldbank.org/indicator/NV.IND.MANF.ZS},
  note = {Series {NV.IND.MANF.ZS}, {World} aggregate; 2025 observation (latest
          available), 14.9\%. Accessed: 2026-04-30}
}

@inproceedings{brown2020language,
  title={Language models are few-shot learners},
  author={Brown, Tom and Mann, Benjamin and Ryder, Nick and Subbiah, Melanie and Kaplan, Jared and Dhariwal, Prafulla and Neelakantan, Arvind and Shyam, Pranav and Sastry, Girish and Askell, Amanda and others},
  booktitle={Advances in Neural Information Processing Systems},
  year={2020},
  url={https://proceedings.neurips.cc/paper/2020/hash/1457c0d6bfcb4967418bfb8ac142f64a-Abstract.html}
}

@inproceedings{dosovitskiy2020image,
  title={An Image is Worth 16x16 Words: Transformers for Image Recognition at Scale},
  author={Dosovitskiy, Alexey and Beyer, Lucas and Kolesnikov, Alexander and Weissenborn, Dirk and Zhai, Xiaohua and Unterthiner, Thomas and Dehghani, Mostafa and Minderer, Matthias and Heigold, Georg and Gelly, Sylvain and Uszkoreit, Jakob and Houlsby, Neil},
  booktitle={International Conference on Learning Representations},
  year={2021},
  url={https://openreview.net/forum?id=YicbFdNTTy}
}

@misc{skab,
  author = {Katser, Iurii D. and Kozitsin, Vyacheslav O.},
  title = {Skoltech Anomaly Benchmark ({SKAB})},
  year = {2020},
  publisher = {Kaggle},
  howpublished = {\url{https://www.kaggle.com/dsv/1693952}},
  doi = {10.34740/KAGGLE/DSV/1693952}
}

@article{veloso2022benchmarkdatasetpredictivemaintenance,
  author  = {Veloso, Bruno and Ribeiro, Rita P. and Gama, Jo{\~a}o and Pereira, Pedro Mota},
  title   = {The {MetroPT} dataset for predictive maintenance},
  journal = {Scientific Data},
  volume  = {9},
  number  = {1},
  pages   = {764},
  year    = {2022},
  doi     = {10.1038/s41597-022-01877-3}
}

@inproceedings{WADI2017testbed,
  author = {Ahmed, Chuadhry Mujeeb and Palleti, Venkata Reddy and Mathur, Aditya P.},
  title = {{WADI}: a water distribution testbed for research in the design of secure cyber physical systems},
  year = {2017},
  publisher = {Association for Computing Machinery},
  address = {New York, NY, USA},
  url = {https://doi.org/10.1145/3055366.3055375},
  doi = {10.1145/3055366.3055375},
  booktitle = {Proceedings of the 3rd International Workshop on Cyber-Physical Systems for Smart Water Networks},
  pages = {25--28},
  numpages = {4}
}

@article{downs1993plant,
  title={A plant-wide industrial process control problem},
  author={Downs, J.J. and Vogel, E.F.},
  journal={Computers \& Chemical Engineering},
  volume={17},
  number={3},
  pages={245--255},
  year={1993},
  publisher={Elsevier},
  doi={10.1016/0098-1354(93)80018-I},
  url={https://doi.org/10.1016/0098-1354(93)80018-I}
}

@inproceedings{liu2024elephant,
  title={The Elephant in the Room: Towards A Reliable Time-Series Anomaly Detection Benchmark},
  author={Liu, Qinghua and Paparrizos, John},
  booktitle={Advances in Neural Information Processing Systems},
  year={2024},
  note={Datasets and Benchmarks Track}
}

@inproceedings{lai2025todsautomatedtimeseries,
  title     = {{TODS}: An Automated Time Series Outlier Detection System},
  author    = {Lai, Kwei-Herng and Zha, Daochen and Wang, Guanchu and Xu, Junjie and Zhao, Yue and Kumar, Devesh and Chen, Yile and Zumkhawaka, Purav and Wan, Minyang and Martinez, Diego and Hu, Xia},
  booktitle = {Proceedings of the AAAI Conference on Artificial Intelligence},
  volume    = {35},
  pages     = {16060--16062},
  year      = {2021}
}

@inproceedings{
xu2022anomaly,
title={Anomaly {T}ransformer: Time Series Anomaly Detection with Association Discrepancy},
author={Jiehui Xu and Haixu Wu and Jianmin Wang and Mingsheng Long},
booktitle={International Conference on Learning Representations},
year={2022},
url={https://openreview.net/forum?id=LzQQ89U1qm_}
}

@article{RAISSI2019686,
title = {Physics-informed neural networks: A deep learning framework for solving forward and inverse problems involving nonlinear partial differential equations},
journal = {Journal of Computational Physics},
volume = {378},
pages = {686-707},
year = {2019},
issn = {0021-9991},
doi = {10.1016/j.jcp.2018.10.045},
url = {https://doi.org/10.1016/j.jcp.2018.10.045},
author = {M. Raissi and P. Perdikaris and G.E. Karniadakis},
}

@article{zeng2020tossingbot,
  title={{TossingBot}: Learning to Throw Arbitrary Objects with Residual Physics},
  author={Zeng, Andy and Song, Shuran and Lee, Johnny and Rodriguez, Alberto and Funkhouser, Thomas},
  journal={{IEEE} Transactions on Robotics},
  volume={36},
  number={4},
  pages={1307--1319},
  year={2020},
  doi={10.1109/TRO.2020.2988642}
}

@misc{sharon_2017_toolwear,
  author       = {Sun, Sharon},
  title        = {{CNC} {M}ill {T}ool {W}ear},
  year         = {2018},
  publisher    = {Kaggle},
  url          = {https://www.kaggle.com/datasets/shasun/tool-wear-detection-in-cnc-mill},
  note         = {CC0 Public Domain; machining experiments at the System-level Manufacturing and Automation Research Testbed (SMART), University of Michigan; dataset metadata last modified 2018-04-06 on Kaggle}
}

@inproceedings{dempster2021minirocket,
  title     = {{MiniRocket}: A Very Fast (Almost) Deterministic Transform for Time Series Classification},
  author    = {Dempster, Angus and Schmidt, Daniel F. and Webb, Geoffrey I.},
  booktitle = {Proceedings of the 27th ACM SIGKDD Conference on Knowledge Discovery \& Data Mining},
  pages     = {248--257},
  year      = {2021},
  publisher = {ACM},
  doi       = {10.1145/3447548.3467231}
}

@inproceedings{sun2016coral,
  title     = {Return of Frustratingly Easy Domain Adaptation},
  author    = {Sun, Baochen and Feng, Jiashi and Saenko, Kate},
  booktitle = {Proceedings of the AAAI Conference on Artificial Intelligence},
  volume    = {30},
  year      = {2016},
  doi       = {10.1609/aaai.v30i1.10306}
}

@article{ganin2016dann,
  title   = {Domain-Adversarial Training of Neural Networks},
  author  = {Ganin, Yaroslav and Ustinova, Evgeniya and Ajakan, Hana and Germain, Pascal and Larochelle, Hugo and Laviolette, Fran{\c{c}}ois and Marchand, Mario and Lempitsky, Victor},
  journal = {Journal of Machine Learning Research},
  volume  = {17},
  number  = {59},
  pages   = {1--35},
  year    = {2016}
}
\bibliographystyle{iclr2027_conference}

\newpage
\appendix
\section{Datasheet for Datasets (Gebru et al., 2021)}
\label{app:datasheet}

Following the recommendations of \citet{gebru2021datasheets}, we provide
a structured datasheet for FactoryNet v1.0 (\cref{tab:datasheet}).

{\footnotesize
\setlength{\tabcolsep}{4pt}%
\begin{longtable}{p{3.3cm}p{10.1cm}}
    \caption{Datasheet for FactoryNet v1.0.}\label{tab:datasheet}\\
    \toprule
    \textbf{Question} & \textbf{Answer} \\
    \midrule
    \endfirsthead
    \toprule
    \textbf{Question} & \textbf{Answer} \\
    \midrule
    \endhead
        \multicolumn{2}{l}{\textit{Motivation}}                                                                                                                                                               \\
        \midrule
        Purpose
                          & Pretraining substrate for industrial time-series foundation
        models, anomaly detection benchmarking, and cross-embodiment
        transfer research.                                                                                                                                                                                    \\
        Creators
                          & Jonas Petersen (ETH~Zurich / Forgis), Pratik Jawahar (Forgis), Riccardo Maggioni (Forgis), Alexandre Huou (EPFL), Karim Othman (Cairo University / Forgis), Camilla Mazzoleni (Forgis), Matteo Cozzi (Forgis), Marek Gajewski (Forgis), Davide Paloschi (ETH~Zurich), Gian-Alessandro Lombardi (Forgis), Matei C. Ignuta-Ciuncanu (UC~Berkeley), Lior Porath (ETH~Zurich), Federico Martelli (ETH~Zurich / Forgis), Philipp Christian Petersen (University of Vienna).                                                                                                                                                              \\
        Funding
                          & Fully funded by Forgis AG.                                                                                                                                                 \\
        \midrule
        \multicolumn{2}{l}{\textit{Composition}}                                                                                                                                                              \\
        \midrule
        Instances
                          & Episodes of robotic task execution, each containing
        multi-channel time-series at the controller rate with S-E-F-C role
        annotations.                                                                                                                                                                                          \\
        Count             & 56,591 episodes: 8,863 real lab (UR3), 1,799 real lab (KUKA KR10), 4,185 open-source (voraus-AD 2,122; AURSAD (UR3e) 2,045; UMich CNC 18), and 41,744 synthetic (Isaac Sim, five arms). 113M logged time steps. 1,553 real counterfactual episodes come in addition. \\
        Sampling
                          & Lab data: programmatic task execution with configurable
        fault injection; of the 10,662 laboratory episodes, 28\% are healthy and 72\% contain an injected fault.
        Open-source: full datasets ingested.
        Synthetic: procedural generation with domain randomization.                                                                                                                                           \\
        Missing data
                          & Sensor dropouts $<$0.1\% of time steps; handled via
        linear interpolation.  No episodes removed due to missing
        data.                                                                                                                                                                                                 \\
        Confidentiality
                          & No PII. Industrial parameters only (joint angles, torques,
        velocities).                                                                                                                                                                                          \\
        \midrule
        \multicolumn{2}{l}{\textit{Collection Process}}                                                                                                                                                       \\
        \midrule
        Acquisition
                          & Lab episodes recorded by the authors via robot controller
        APIs (RTDE for UR3, RSI for KUKA KR10). Synthetic episodes are logged at 125~Hz (UR) and 83.3~Hz (KR10, the KRC interpolation cycle), with no resampling.
        Open-source subsets downloaded and re-processed into S-E-F-C
        format via adapter scripts.                                                                                                                                                                           \\
        Mechanisms
                          & Automated scripts executing task programs with fault
        injection at configurable rates.  27 anomaly types, prepared physically on the cell where the fault requires it (for example a loosened fixture or an added axis weight) and otherwise injected
        programmatically (payload variation, misgrips, collisions,
        stripped threads, jamming, friction changes, etc.).                                                                                                                                                   \\
        Who collected
                          & Yanis Merzouki, Karim Othman, Balazs Gunther.                                                                             \\
        Time frame
                          & March 1 -- April 20, 2026.                                                                                                                                                        \\
        Ethical review
                          & N/A: no human subjects.                                                                                                                                                        \\
        \midrule
        \multicolumn{2}{l}{\textit{Preprocessing, Cleaning, and Labeling}}                                                                                                                                    \\
        \midrule
        Preprocessing
                          & NaN interpolation and S-E-F-C column mapping per source
        via adapter scripts; each source keeps its native controller rate.  No
        per-channel normalisation applied (raw physical units preserved).                                                                                                                                                 \\
        Raw data
                          & Raw recordings available on request.  Distributed format:
        S-E-F-C Parquet files with metadata.                                                                                                                                                                  \\
        Labeling
                          & Programmatic fault injection with known ground truth; no
        manual annotation required for lab data.  Open-source
        subsets retain original labels, mapped to unified anomaly
        taxonomy.                                                                                                                                                                                             \\
        \midrule
        \multicolumn{2}{l}{\textit{Uses}}                                                                                                                                                                     \\
        \midrule
        Prior uses
                          & Anomaly detection baselines (this paper), cross-embodiment
        transfer experiments (this paper).                                                                                                                                                                    \\
        Other uses
                          & Forecasting, remaining useful life estimation, control
        policy learning, sim-to-real benchmarking, counterfactual
        analysis.                                                                                                                                                                                             \\
        Misuse potential
                          & Dataset reflects specific robot models and lab conditions;
        generalisation claims to unseen embodiments or industrial
        deployments should be scoped accordingly.                                                                                                                                                             \\
        \midrule
        \multicolumn{2}{l}{\textit{Distribution}}                                                                                                                                                             \\
        \midrule
        Distribution
                          & Distributed as S-E-F-C Parquet files with metadata and
        framework-native dataloaders; the public copy is linked from
        the first page of this paper.                                                                                                                                                           \\
        License
                          & Novel lab and synthetic data under MIT. Open-source subsets retain their
        original licenses and are redistributed under them: voraus-AD is
        CC~BY-NC-SA~4.0 (NonCommercial, ShareAlike), AURSAD is CC~BY~4.0, and the
        UMich CNC subset follows its source terms.                                                                                                                                                   \\
        Restrictions
                          & The voraus-AD subset is CC~BY-NC-SA~4.0: NonCommercial and
        ShareAlike. Commercial use of that subset, and redistribution of works derived
        from it under more permissive terms, are not permitted. Novel and synthetic data
        carry no such restriction. No export controls apply.                            \\
        \midrule
        \multicolumn{2}{l}{\textit{Maintenance}}                                                                                                                                                              \\
        \midrule
        Maintainer
                          & Forgis AG.
        Issue tracker: GitHub repository.                                                                                                                                                                    \\
        Updates
                          & Semantic versioning (v1.0 at submission).  New embodiments
        and tasks increment minor version; schema changes increment
        major version.  Prior versions remain available.                                                                                                                                                      \\
        Retention
                          & All released versions are retained permanently.                                                                                                                                   \\
        Errata
                          & Reported and tracked in the public issue tracker of the released repository; corrections published as patch versions.                                                                                                                                                                          \\
        \bottomrule
\end{longtable}
}

\section{Signal-to-S-E-F-C Mapping}
\label{app:sefc_mapping}

We provide one table per source mapping raw signal names to S-E-F-C roles.
UR3, KUKA KR10 and Isaac Sim mappings are listed with raw signal names confirmed;
CNC, AURSAD, and voraus-AD mappings are fully
confirmed.

\subsection*{B.1\quad UR3 Laboratory Source}
Signals of UR3 can be found at Table ~\ref{tab:sefc_mapping_ur3e}

\begin{table}[!p]
    \centering
    \caption{S-E-F-C mapping for the UR3 laboratory source.}
    \label{tab:sefc_mapping_ur3e}
    \footnotesize
    \setlength{\tabcolsep}{4pt}%
    \begin{tabular}{@{}lll>{\raggedright\arraybackslash}p{2.4cm}@{}}
        \toprule
        \textbf{Raw Signal Name}                                & \textbf{S-E-F-C Role} & \textbf{Unit} & \textbf{Notes}              \\
        \midrule
        \texttt{machine\_id}                                    & Metadata              & --           & Episode metadata            \\
        \texttt{ur3\_robot\_target\_joint\_0\ldots5}            & Setpoint              & rad           & 6 joints                    \\
        \texttt{ur3\_robot\_target\_joint\_vel\_0\ldots5}       & Setpoint              & rad/s         & 6 joints                    \\
        \texttt{ur3\_robot\_joint\_0\ldots5}                    & Feedback              & rad           & Actual positions, 6 joints  \\
        \texttt{ur3\_robot\_joint\_vel\_0\ldots5}               & Feedback              & rad/s         & Actual velocities, 6 joints \\
        \texttt{ur3\_robot\_joint\_current\_0\ldots5}           & Effort                & A             & 6 joints                    \\
        \texttt{ur3\_robot\_target\_joint\_current\_0\ldots5}   & Effort                & A             & Commanded current, 6 joints \\
        \texttt{ur3\_robot\_joint\_temp\_0\ldots5}              & Context               & \textdegree C & 6 joints                    \\
        \texttt{ur3\_robot\_joint\_control\_output\_0\ldots5}   & Effort                & --           & 6 joints                    \\
        \texttt{ur3\_robot\_joint\_mode\_0\ldots5}              & Context               & enum          & 6 joints                    \\
        \texttt{ur3\_robot\_tcp\_x/y/z}                         & Feedback              & m             & Actual TCP position         \\
        \texttt{ur3\_robot\_tcp\_rx/ry/rz}                      & Feedback              & rad           & Actual TCP orientation      \\
        \texttt{ur3\_robot\_target\_tcp\_x/y/z}                 & Setpoint              & m             & Target TCP position         \\
        \texttt{ur3\_robot\_target\_tcp\_rx/ry/rz}              & Setpoint              & rad           & Target TCP orientation      \\
        \texttt{ur3\_robot\_tcp\_speed\_x/y/z/rx/ry/rz}         & Feedback              & m/s \& rad/s  & Actual TCP speed (6 axes)   \\
        \texttt{ur3\_robot\_target\_tcp\_speed\_x/y/z/rx/ry/rz} & Setpoint              & m/s \& rad/s  & Target TCP speed (6 axes)   \\
        \texttt{ur3\_robot\_tcp\_force\_x/y/z}                  & Effort                & N             & TCP force                   \\
        \texttt{ur3\_robot\_tcp\_torque\_x/y/z}                 & Effort                & Nm            & TCP torque                  \\
        \texttt{ur3\_robot\_robot\_mode}                        & Context               & enum          & Enum                        \\
        \texttt{ur3\_robot\_safety\_mode}                       & Context               & enum          & Enum                        \\
        \texttt{ur3\_robot\_digital\_inputs}                    & Context               & bitmask       & Bitmask                     \\
        \texttt{ur3\_robot\_digital\_outputs}                   & Context               & bitmask       & Bitmask                     \\
        \texttt{ur3\_robot\_robot\_status}                      & Context               & enum          &                             \\
        \texttt{ur3\_robot\_safety\_status}                     & Context               & enum          &                             \\
        \texttt{ur3\_robot\_runtime\_state}                     & Context               & enum          &                             \\
        \texttt{ur3\_robot\_main\_voltage}                      & Context               & V             & V                           \\
        \texttt{ur3\_robot\_robot\_voltage}                     & Context               & V             & V                           \\
        \texttt{ur3\_robot\_robot\_current}                     & Context               & A             & A                           \\
        \texttt{ur3\_robot\_analog\_input\_0/1}                 & Context               & V/A           &                             \\
        \texttt{ur3\_robot\_analog\_output\_0/1}                & Context               & V/A           &                             \\
        \texttt{ur3\_robot\_speed\_scaling}                     & Context               & \%            &                             \\
        \texttt{ur3\_robot\_target\_speed\_fraction}            & Context               & \%            &                             \\
        \texttt{ur3\_robot\_momentum}                           & Context               & kg$\cdot$m/s  &                             \\
        \texttt{realsense\_camera\_frame\_count}                & Context               & count         &                             \\
        \texttt{realsense\_camera\_connected}                   & Context               & bool          &                             \\
        \texttt{realsense\_camera\_streaming}                   & Context               & bool          &                             \\
        \bottomrule
    \end{tabular}
\end{table}

\subsection*{B.2\quad KUKA KR10 Laboratory Source}
Signals of KUKA KR10 can be found at Table ~\ref{tab:sefc_mapping_kuka}
\begin{table}[!ht]
    \centering
    \caption{S-E-F-C mapping for the KUKA KR10 laboratory source. Note hardware limitations of KSS 8.3.}
    \label{tab:sefc_mapping_kuka}
    \resizebox{\textwidth}{!}{%
        \begin{tabular}{llll}
            \toprule
            \textbf{Signal Group} & \textbf{Described Fields}              & \textbf{S-E-F-C Role} & \textbf{Unit} \\
            \midrule
            Position (actual)     & Actual joint positions ($\times$6)     & Feedback              & degrees       \\
            Position (commanded)  & Commanded joint positions ($\times$6)  & Setpoint              & degrees       \\
            TCP pose (actual)     & Actual TCP pose ($\times$6)            & Feedback              & mm / degrees  \\
            Motor current         & Motor current per axis ($\times$6)     & Effort                & \% of max     \\
            Motor torque          & Motor torque per axis ($\times$6)      & Effort                & Nm            \\
            Motor temperature     & Motor temperature per axis ($\times$6) & Context               & Kelvin        \\
            Cartesian accel.      & Accel.\ X, Y, Z, abs ($\times$4)       & Feedback              & m/s$^2$       \\
            Speed override        & Speed override (scalar)                & Context               & \%            \\
            Process state         & State enum (FREE/ACTIVE/STOP/END)      & Context               & enum          \\
            Digital I/O           & Digital inputs bitmask                 & Context               & bitmask       \\
            Digital I/O           & Digital outputs bitmask                & Context               & bitmask       \\
            \midrule
            \multicolumn{4}{l}{\textit{Not available on KSS 8.3: joint velocities, commanded TCP
                    pose, TCP force/torque, joint voltage, tool accelerometer, joint control output.}}             \\
            \bottomrule
        \end{tabular}%
    }
\end{table}

\subsection*{B.3\quad UMich CNC Source}
Mapping of CNC columns (Table \ref{tab:sefc_mapping_cnc})
\begin{table}[!ht]
    \centering
    \caption{S-E-F-C mapping for the UMich CNC milling source.
        Axes: X1, Y1, Z1 (linear), S1 (spindle).}
    \label{tab:sefc_mapping_cnc}
    \resizebox{\textwidth}{!}{%
        \begin{tabular}{lllll}
            \toprule
            \textbf{Raw Signal Name}              & \textbf{S-E-F-C Name}          & \textbf{Role}
                                                  & \textbf{Unit}                  & \textbf{Axis}                      \\
            \midrule
            \texttt{X1\_CommandPosition}          & \texttt{setpoint\_pos\_0}      & Setpoint      & mm       & X       \\
            \texttt{X1\_CommandVelocity}          & \texttt{setpoint\_vel\_0}      & Setpoint      & mm/s     & X       \\
            \texttt{X1\_CommandAcceleration}      & \texttt{setpoint\_acc\_0}      & Setpoint      & mm/s$^2$ & X       \\
            \texttt{X1\_ActualPosition}           & \texttt{feedback\_pos\_0}      & Feedback      & mm       & X       \\
            \texttt{X1\_ActualVelocity}           & \texttt{feedback\_vel\_0}      & Feedback      & mm/s     & X       \\
            \texttt{X1\_ActualAcceleration}       & \texttt{feedback\_acc\_0}      & Feedback      & mm/s$^2$ & X       \\
            \texttt{X1\_CurrentFeedback}          & \texttt{feedback\_current\_0}  & Feedback      & A        & X       \\
            \texttt{X1\_OutputCurrent}            & \texttt{effort\_current\_0}    & Effort        & A        & X       \\
            \texttt{X1\_OutputVoltage}            & \texttt{effort\_voltage\_0}    & Effort        & V        & X       \\
            \texttt{X1\_OutputPower}              & \texttt{effort\_power\_0}      & Effort        & W        & X       \\
            \texttt{X1\_DCBusVoltage}             & \texttt{ctx\_busvoltage\_0}    & Context       & V        & X       \\
            \multicolumn{5}{l}{\textit{(Pattern repeats for Y1 $\to$ axis 1, Z1 $\to$ axis 2)}}                         \\
            \midrule
            \texttt{S1\_CommandPosition}          & \texttt{setpoint\_pos\_3}      & Setpoint      & deg      & Spindle \\
            \texttt{S1\_CommandVelocity}          & \texttt{setpoint\_vel\_3}      & Setpoint      & rpm      & Spindle \\
            \texttt{S1\_CommandAcceleration}      & \texttt{setpoint\_acc\_3}      & Setpoint      & rpm/s    & Spindle \\
            \texttt{S1\_ActualPosition}           & \texttt{feedback\_pos\_3}      & Feedback      & deg      & Spindle \\
            \texttt{S1\_ActualVelocity}           & \texttt{feedback\_vel\_3}      & Feedback      & rpm      & Spindle \\
            \texttt{S1\_ActualAcceleration}       & \texttt{feedback\_acc\_3}      & Feedback      & rpm/s    & Spindle \\
            \texttt{S1\_CurrentFeedback}          & \texttt{feedback\_current\_3}  & Feedback      & A        & Spindle \\
            \texttt{S1\_OutputCurrent}            & \texttt{effort\_current\_3}    & Effort        & A        & Spindle \\
            \texttt{S1\_OutputVoltage}            & \texttt{effort\_voltage\_3}    & Effort        & V        & Spindle \\
            \texttt{S1\_OutputPower}              & \texttt{effort\_power\_3}      & Effort        & W        & Spindle \\
            \texttt{S1\_DCBusVoltage}             & \texttt{ctx\_busvoltage\_3}    & Context       & V        & Spindle \\
            \texttt{S1\_SystemInertia}            & \texttt{ctx\_inertia\_3}       & Context       & --      & Spindle \\
            \midrule
            \texttt{M1\_CURRENT\_PROGRAM\_NUMBER} & \texttt{ctx\_program\_number}  & Context       & --      & --     \\
            \texttt{M1\_sequence\_number}         & \texttt{ctx\_sequence\_number} & Context       & --      & --     \\
            \texttt{M1\_CURRENT\_FEEDRATE}        & \texttt{ctx\_feedrate}         & Context       & mm/min   & --     \\
            \texttt{Machining\_Process}           & \texttt{ctx\_process\_phase}   & Context       & enum     & --     \\
            \bottomrule
        \end{tabular}%
    }
\end{table}

\subsection*{B.4\quad AURSAD Source (UR3e Screwdriving)}

Per-joint signals follow a fixed pattern for joints $i \in \{0,\ldots,5\}$;
one representative joint (joint~0) is shown below.
(Table~\ref{tab:sefc_mapping_aursad})
\begin{table}[!ht]
    \centering
    \caption{S-E-F-C mapping for the AURSAD source (representative
        joint~0 pattern; repeats for joints 1--5 with incremented index).}
    \label{tab:sefc_mapping_aursad}
    \resizebox{\textwidth}{!}{%
        \begin{tabular}{lllll}
            \toprule
            \textbf{Raw Signal Name}                    & \textbf{S-E-F-C Name}                     & \textbf{Role}
                                                        & \textbf{Unit}                             & \textbf{Notes}                                                  \\
            \midrule
            \texttt{target\_q\_0}                       & \texttt{setpoint\_pos\_0}                 & Setpoint       & rad           & Joint 0                        \\
            \texttt{target\_qd\_0}                      & \texttt{setpoint\_vel\_0}                 & Setpoint       & rad/s         &                                \\
            \texttt{target\_qdd\_0}                     & \texttt{setpoint\_acc\_0}                 & Setpoint       & rad/s$^2$     &                                \\
            \texttt{target\_current\_0}                 & \texttt{setpoint\_current\_0}             & Setpoint       & A             &                                \\
            \texttt{target\_moment\_0}                  & \texttt{setpoint\_torque\_0}              & Setpoint       & Nm            &                                \\
            \texttt{actual\_q\_0}                       & \texttt{feedback\_pos\_0}                 & Feedback       & rad           &                                \\
            \texttt{actual\_qd\_0}                      & \texttt{feedback\_vel\_0}                 & Feedback       & rad/s         &                                \\
            \texttt{actual\_current\_0}                 & \texttt{effort\_current\_0}               & Effort         & A             &                                \\
            \texttt{actual\_control\_output\_0}         & \texttt{effort\_control\_0}               & Effort         & --           &                                \\
            \texttt{actual\_joint\_voltage\_0}          & \texttt{effort\_voltage\_0}               & Effort         & V             &                                \\
            \texttt{joint\_temperatures\_0}             & \texttt{ctx\_temp\_0}                     & Context        & \textdegree C &                                \\
            \texttt{joint\_mode\_0}                     & \texttt{ctx\_joint\_mode\_0}              & Context        & enum          &                                \\
            \texttt{target\_TCP\_pose\_0}               & \texttt{setpoint\_pos\_cartesian\_0}      & Setpoint       & m/rad         &                                \\
            \texttt{target\_TCP\_speed\_0}              & \texttt{setpoint\_vel\_cartesian\_0}      & Setpoint       & m/s           &                                \\
            \texttt{actual\_TCP\_pose\_0}               & \texttt{feedback\_pos\_cartesian\_0}      & Feedback       & m/rad         &                                \\
            \texttt{actual\_TCP\_speed\_0}              & \texttt{feedback\_vel\_cartesian\_0}      & Feedback       & m/s           &                                \\
            \texttt{actual\_TCP\_force\_0}              & \texttt{effort\_force\_cartesian\_0}      & Effort         & N             &                                \\
            \texttt{actual\_tool\_accelerometer\_0/1/2} & \texttt{auxiliary\_accel\_tool\_0/1/2}
                                                        & --                                       & m/s$^2$        & Tool IMU                                       \\
            \midrule
            \texttt{output\_double\_register\_24}       & \texttt{feedback\_torque\_tool}           & Feedback       & Nm            & Tool torque                    \\
            \texttt{output\_double\_register\_25}       & \texttt{effort\_torque\_tool}             & Effort         & Nm            &                                \\
            \texttt{output\_double\_register\_26}       & \texttt{setpoint\_torque\_tool}           & Setpoint       & Nm            &                                \\
            \texttt{output\_double\_register\_27}       & \texttt{setpoint\_torque\_gradient\_tool} & Setpoint       & Nm/s          &                                \\
            \midrule
            \texttt{actual\_main\_voltage}              & \texttt{ctx\_main\_voltage}               & Context        & V             &                                \\
            \texttt{actual\_robot\_voltage}             & \texttt{ctx\_robot\_voltage}              & Context        & V             &                                \\
            \texttt{actual\_robot\_current}             & \texttt{ctx\_robot\_current}              & Context        & A             &                                \\
            \texttt{speed\_scaling}                     & \texttt{ctx\_speed\_scaling}              & Context        & --           &                                \\
            \texttt{target\_speed\_fraction}            & \texttt{ctx\_target\_speed\_fraction}     & Context        & --           &                                \\
            \texttt{actual\_momentum}                   & \texttt{ctx\_momentum}                    & Context        & kg$\cdot$m/s  &                                \\
            \texttt{robot\_mode}                        & \texttt{ctx\_robot\_mode}                 & Context        & enum          &                                \\
            \texttt{safety\_mode}                       & \texttt{ctx\_safety\_mode}                & Context        & enum          &                                \\
            \texttt{runtime\_state}                     & \texttt{ctx\_runtime\_state}              & Context        & enum          &                                \\
            \texttt{label}                              & \texttt{raw\_label}                       & --            & --           & 0=Normal, 1=Damaged screw,     \\
                                                        &                                           &                &               & 2=Extra part, 3=Missing screw, \\
                                                        &                                           &                &               & 4=Damaged thread               \\
            \bottomrule
        \end{tabular}%
    }
\end{table}

\subsection*{B.5\quad voraus-AD Source (Yu-Cobot)}

Per-joint signals follow the same index pattern for joints
$i \in \{1,\ldots,6\}$ (1-indexed in source); one representative joint
is shown. (Table \ref{tab:sefc_mapping_voraus})

\begin{table}[!ht]
    \centering
    \caption{S-E-F-C mapping for the voraus-AD source (representative
        joint~1 pattern; repeats for joints 2--6).}
    \label{tab:sefc_mapping_voraus}
    \small
    \begin{tabular}{lllll}
        \toprule
        \textbf{Raw Signal Name}         & \textbf{S-E-F-C Name}             & \textbf{Role}
                                         & \textbf{Unit}                     & \textbf{Notes}                                 \\
        \midrule
        \texttt{target\_position\_1}     & \texttt{setpoint\_pos\_0}         & Setpoint       & rad       & Joint 1$\to$idx 0 \\
        \texttt{target\_velocity\_1}     & \texttt{setpoint\_vel\_0}         & Setpoint       & rad/s     &                   \\
        \texttt{target\_acceleration\_1} & \texttt{setpoint\_acc\_0}         & Setpoint       & rad/s$^2$ &                   \\
        \texttt{target\_torque\_1}       & \texttt{setpoint\_torque\_0}      & Setpoint       & Nm        &                   \\
        \texttt{joint\_position\_1}      & \texttt{feedback\_pos\_0}         & Feedback       & rad       &                   \\
        \texttt{joint\_velocity\_1}      & \texttt{feedback\_vel\_0}         & Feedback       & rad/s     &                   \\
        \texttt{motor\_position\_1}      & \texttt{feedback\_motor\_pos\_0}  & Feedback       & rad       &                   \\
        \texttt{motor\_velocity\_1}      & \texttt{feedback\_motor\_vel\_0}  & Feedback       & rad/s     &                   \\
        \texttt{torque\_sensor\_a\_1}    & \texttt{feedback\_torque\_a\_0}   & Feedback       & Nm        &                   \\
        \texttt{torque\_sensor\_b\_1}    & \texttt{feedback\_torque\_b\_0}   & Feedback       & Nm        &                   \\
        \texttt{motor\_torque\_1}        & \texttt{effort\_motor\_torque\_0} & Effort         & Nm        &                   \\
        \texttt{motor\_iq\_1}            & \texttt{effort\_current\_iq\_0}   & Effort         & A         &                   \\
        \texttt{motor\_id\_1}            & \texttt{effort\_current\_id\_0}   & Effort         & A         &                   \\
        \texttt{power\_motor\_el\_1}     & \texttt{effort\_power\_el\_0}     & Effort         & W         &                   \\
        \texttt{power\_motor\_mech\_1}   & \texttt{effort\_power\_mech\_0}   & Effort         & W         &                   \\
        \texttt{power\_load\_mech\_1}    & \texttt{effort\_power\_load\_0}   & Effort         & W         &                   \\
        \texttt{motor\_voltage\_1}       & \texttt{effort\_voltage\_0}       & Effort         & V         &                   \\
        \texttt{computed\_inertia\_1}    & \texttt{ctx\_inertia\_0}          & Context        & --       &                   \\
        \texttt{computed\_torque\_1}     & \texttt{ctx\_computed\_torque\_0} & Context        & Nm        &                   \\
        \texttt{supply\_voltage\_1}      & \texttt{ctx\_busvoltage\_0}       & Context        & V         &                   \\
        \texttt{brake\_voltage\_1}       & \texttt{ctx\_brake\_voltage\_0}   & Context        & V         &                   \\
        \midrule
        \texttt{robot\_voltage}          & \texttt{ctx\_robot\_voltage}      & Context        & V         & Global            \\
        \texttt{robot\_current}          & \texttt{ctx\_robot\_current}      & Context        & A         & Global            \\
        \texttt{io\_current}             & \texttt{ctx\_io\_current}         & Context        & A         & Global            \\
        \texttt{system\_current}         & \texttt{ctx\_system\_current}     & Context        & A         & Global            \\
        \texttt{anomaly}                 & \texttt{ctx\_is\_anomaly}         & --            & bool      & Label             \\
        \texttt{category}                & \texttt{ctx\_anomaly\_category}   & --            & str       & Fault type        \\
        \texttt{setting}                 & \texttt{ctx\_setting}             & --            & str       &                   \\
        \bottomrule
    \end{tabular}
\end{table}

\subsection*{B.6\quad Isaac Sim UR5 (Synthetic)}
Mapping of Isaac Sim UR5 columns (Table ~\ref{tab:sefc_mapping_isaac})
\begin{table}[!ht]
    \centering
    \caption{S-E-F-C mapping for the earlier Isaac Sim (UR5) synthetic source, retained for completeness: signals are procedurally generated and logged at a base 60~Hz simulation step before interpolation. This is \emph{not} the campaign behind the results reported in the main text, which runs at 500~Hz physics and logs at 125~Hz (UR) and 83.3~Hz (KR10) with no resampling (Appendix~\ref{app:sim_physics}).}
    \label{tab:sefc_mapping_isaac}
    \resizebox{\textwidth}{!}{%
        \begin{tabular}{llll}
            \toprule
            \textbf{Signal Group (Raw Prefix)}             & \textbf{Described Fields}              & \textbf{S-E-F-C Role} & \textbf{Unit}       \\
            \midrule
            \texttt{episode}, \texttt{step}, \texttt{time} & Global/episode steps, sim/wall time    & Metadata              & count / s           \\
            \texttt{state\_machine}, \texttt{phase}        & Task, controller phases, event IDs     & Context               & enum / str          \\
            \midrule
            \texttt{joint\_cmd\_pos\_rad\_*}               & Commanded joint positions ($\times$6)  & Setpoint              & rad                 \\
            \texttt{joint\_cmd\_vel\_radps\_*}             & Commanded joint velocities ($\times$6) & Setpoint              & rad/s               \\
            \texttt{joint\_pos\_rad\_*}                    & Actual joint positions ($\times$6)     & Feedback              & rad                 \\
            \texttt{joint\_vel\_radps\_*}                  & Actual joint velocities ($\times$6)    & Feedback              & rad/s               \\
            \texttt{joint\_accel\_radps2\_*}               & Actual joint accelerations ($\times$6) & Feedback              & rad/s$^2$           \\
            \texttt{joint\_torque\_nm\_*}                  & Joint applied torques ($\times$6)      & Effort                & Nm                  \\
            \texttt{joint\_pos\_error\_rad\_*}             & Tracking error per joint ($\times$6)   & Context               & rad                 \\
            \midrule
            \texttt{ee\_cmd\_pos\_*}, \texttt{quat\_*}     & Commanded TCP pose (pos, quat)         & Setpoint              & m / --             \\
            \texttt{ee\_pos\_*}, \texttt{quat\_*}          & Actual TCP pose (pos, quat)            & Feedback              & m / --             \\
            \texttt{ee\_euler\_*\_rad}                     & Actual TCP orientation (RPY)           & Feedback              & rad                 \\
            \texttt{ee\_linvel\_*}, \texttt{angvel\_*}     & TCP velocities (linear, angular)       & Feedback              & m/s / rad/s         \\
            \texttt{ee\_linacc\_*}, \texttt{angacc\_*}     & TCP accelerations (linear, angular)    & Feedback              & m/s$^2$ / rad/s$^2$ \\
            \midrule
            \texttt{gripper\_cmd\_rad}                     & Commanded gripper position             & Setpoint              & rad                 \\
            \texttt{gripper\_pos\_rad}                     & Actual gripper position                & Feedback              & rad                 \\
            \texttt{gripper\_attached}                     & Gripper logical contact state          & Context               & bool                \\
            \texttt{contact\_force\_*\_n}                  & Physical EE contact force (XYZ, mag)   & Effort                & N                   \\
            \texttt{contact\_torque\_*\_nm}                & Physical EE contact torque (XYZ, mag)  & Effort                & Nm                  \\
            \midrule
            \texttt{cube\_pos\_*}, \texttt{quat\_*}        & Target object pose (pos, quat)         & Context               & m / --             \\
            \texttt{cube\_linvel\_*}, \texttt{angvel\_*}   & Target object velocities               & Context               & m/s / rad/s         \\
            \texttt{ee\_cube\_offset\_*}                   & Relative distance to target            & Context               & m                   \\
            \midrule
            \texttt{cube\_mass\_kg}                        & Domain rand: payload mass              & Context               & kg                  \\
            \texttt{cube\_friction\_coeff}                 & Domain rand: target object friction    & Context               & --                 \\
            \texttt{cube\_width/depth/height\_m}           & Domain rand: target dimensions         & Context               & m                   \\
            \bottomrule
        \end{tabular}%
    }
\end{table}

\section{Full Fault Catalog}
\label{app:fault_catalog}
Table~\ref{tab:fault_catalog} lists all 27 annotated anomaly types used during laboratory data collection. One of the 27, the loosening phase, is a normal phase with a distinct signature rather than a fault; it is counted as an anomaly type to match the AURSAD label schema, so the corpus contains 26 fault types plus that phase.
\begin{table}
    \centering
    \caption{Complete Fault Catalog. Checkmarks ($\checkmark$) indicate whether the fault is included in the Pick-and-Place (PP), Screwing (Scr), and Peg-in-Hole (PiH) task recordings. Crosses ($\times$) indicate absence.}
    \label{tab:fault_catalog}
    \scriptsize
    \setlength{\tabcolsep}{3pt}%
    \begin{tabular}{>{\raggedright\arraybackslash}p{2.4cm} >{\raggedright\arraybackslash}p{5.2cm} >{\raggedright\arraybackslash}p{3.9cm} c c c}
        \toprule
        \textbf{Fault}                  & \textbf{Explanation}                                                                                                                            & \textbf{Injection Procedure}                                                                            & \textbf{PP}  & \textbf{Scr} & \textbf{PiH} \\
        \midrule
        Damaged screw thread            & Screw thread physically damaged, preventing proper engagement during tightening.                                                                & Pre-damaged the screw thread with sandpaper.                                                            & $\times$     & $\checkmark$ & $\times$     \\

        Missing screw                   & Tightening is attempted with no screw present.                                                                                                  & Removed the screw from screwdriver before the cycle.                                                    & $\times$     & $\checkmark$ & $\times$     \\

        Damaged plate thread            & The threaded hole in the plate is damaged, preventing engagement.                                                                               & Pre-damaged the plate hole with a metal screwdriver.                                                    & $\times$     & $\checkmark$ & $\times$     \\

        Loosening phase                 & A counterclockwise loosening rotation replaces tightening; a normal phase with a distinct signal. Counted as an anomaly to match AURSAD schema. & Reversed the rotation direction in the controller program to execute a counterclockwise loosening pass. & $\times$     & $\checkmark$ & $\times$     \\

        Gripper activation failure      & The vacuum gripper fails to activate and never picks up the box.                                                                                & Disabled the gripper activation command in the robot program.                                           & $\checkmark$ & $\times$     & $\times$     \\

        Gripper release during motion   & The gripper releases the payload mid-trajectory, causing an abrupt payload loss.                                                                & Triggered a programmatic gripper-release command at a scripted timestep mid-trajectory.                 & $\checkmark$ & $\times$     & $\times$     \\

        Additional axis payload         & A dead weight attached to one link increases inertia and gravity loading on all joints.                                                         & Bolted calibrated weights of varying mass to one robot link.                                            & $\checkmark$ & $\checkmark$ & $\times$     \\

        Collision with foam object      & Contact with a soft foam block produces a brief TCP force spike without a protective stop.                                                      & Placed a foam cube directly in the programmed TCP trajectory.                                           & $\checkmark$ & $\checkmark$ & $\checkmark$ \\

        Unexpected payload weight       & The transported box has a weight that deviates from the nominal payload configuration.                                                          & Swapped the nominal box for a known heavier or lighter one.                                             & $\checkmark$ & $\times$     & $\times$     \\

        Invalid gripping position       & Timing error: the gripper closes after the arm has begun lifting, gripping the object off-centre.                                               & Added a sleep() call in the controller program so gripper closure lags the lift motion.                 & $\checkmark$ & $\times$     & $\times$     \\

        Unstable mounting platform      & Base instability introduces low-frequency vibrations into the arm.                                                                              & Placed 8 layers of towels under the base plate.                                                         & $\checkmark$ & $\times$     & $\times$     \\

        Joint position limit violation  & A joint moves outside its configured soft or hard position limits, triggering a safety stop.                                                    & Random waypoint slightly beyond the configured soft joint limit.                                        & $\checkmark$ & $\times$     & $\times$     \\

        TCP frame misconfiguration      & TCP frame or mounting orientation misconfigured at the controller; gravity torques deviate.                                                     & Entered an incorrect TCP offset or mounting angle in the controller settings.                           & $\checkmark$ & $\checkmark$ & $\checkmark$ \\

        Payload weight misconfiguration & Payload mass misconfigured while a tool or workpiece is physically attached.                                                                    & Left configured payload weight at multiple wrong mass configurations.                                   & $\checkmark$ & $\checkmark$ & $\times$     \\

        External arm disturbance        & A continuous external force pulls or pushes the TCP during motion.                                                                              & Anchored an elastic rope between the bench frame and the tool flange.                                   & $\checkmark$ & $\checkmark$ & $\checkmark$ \\

        Mild payload CoG misconfig.     & Payload CoG specified at an incorrect offset from the tool flange; gravity compensation is wrong.                                               & Entered a CoG offset in the payload configuration that differs from physical CoG.                       & $\checkmark$ & $\checkmark$ & $\checkmark$ \\

        Collision with hanging cable    & A loose cable drags along a robot link or the tool flange during motion.                                                                        & Suspended a loose cable across the trajectory at arm height.                                            & $\checkmark$ & $\checkmark$ & $\checkmark$ \\

        Collision with cardboard        & A cardboard carton in the trajectory provides moderate resistance; may trigger a protective stop.                                               & Placed a free-standing or braced cardboard box in the programmed trajectory.                            & $\checkmark$ & $\checkmark$ & $\checkmark$ \\

        Collision with rigid object     & A rigid plastic block in the TCP path triggers an immediate protective stop.                                                                    & Clamped a rigid plastic block in the programmed TCP path.                                               & $\checkmark$ & $\checkmark$ & $\checkmark$ \\

        Peg insertion misalignment      & Peg approaches the hole at an angular or lateral offset and jams against the rim.                                                               & Added a recorded lateral offset at the insertion approach waypoint.                                     & $\times$     & $\times$     & $\checkmark$ \\

        Hole obstruction                & Foreign object or debris in the hole prevents full insertion.                                                                                   & Placed a small object (metal screw) inside the hole before insertion.                                   & $\times$     & $\times$     & $\checkmark$ \\

        Incorrect insertion depth       & Insertion terminates at an incorrect Z depth due to a misconfigured waypoint.                                                                   & Shifted the insertion endpoint waypoint Z by recorded offset from nominal.                              & $\times$     & $\times$     & $\checkmark$ \\

        Peg surface contamination       & Contamination or roughness on the peg raises insertion friction and produces stick-slip.                                                        & Applied dry chalk powder to the peg surface.                                                            & $\times$     & $\times$     & $\checkmark$ \\

        Fixture displacement            & Insertion fixture shifted laterally, breaking the match between programmed approach and actual hole.                                            & Physically shifted the hole fixture by recorded offset without updating program.                        & $\times$     & $\times$     & $\checkmark$ \\

        Self-collision                  & Arm configuration causes a link to contact the robot body or mounting fixture.                                                                  & Random waypoint that forces a self-colliding configuration.                                             & $\checkmark$ & $\times$     & $\times$     \\

        Missing box                     & Box absent from the pick position; the gripper closes on empty air.                                                                             & Removed the box from the pick position before the episode started.                                      & $\checkmark$ & $\times$     & $\times$     \\

        Missing peg                     & Arm descends into the hole empty-handed; insertion produces no contact force.                                                                   & Removed the peg from the gripper before the episode started.                                            & $\times$     & $\times$     & $\checkmark$ \\
        \bottomrule
    \end{tabular}
\end{table}

\section{Synthetic Validity Gate and Acceptance Rates}
\label{app:sim_validity}
Fault onsets are drawn uniformly over the phases in which the fault is valid
(payload addition: lift, move, lower; gripper activation failure: settle, close;
gripper release: lift, move; path obstacle: move; miscommutation and collision:
motion phases), with a 30-step margin. A fault episode is kept only if it leaves
a physical trace: an outcome failure or slip, a peak torque above 1.15$\times$ the
nominal envelope, a carry-torque shift above 3 standard deviations, a grip stall
below 0.5$\times$ the nominal minimum, or a protective stop. Path-obstacle
episodes additionally need contact evidence. Table~\ref{tab:sim_validity} lists
valid episodes per arm and condition (ids: 6 payload, 12 miscommutation, 13 gripper
activation failure, 14 gripper release, 16 collision, 17 path obstacle). Rates are out of 2,500 nominal and 1,250 per fault
condition (KR10 path obstacle: 2,500 generated; miscommutation not generated).

\begin{table}[h]
    \centering
    \caption{Valid episodes per arm and condition after the validity gate.}
    \label{tab:sim_validity}
    \small
    \setlength{\tabcolsep}{4pt}%
    \begin{tabular}{lrrrrrrrr}
        \toprule
        Arm & nominal & 6 & 12 & 13 & 14 & 16 & 17 & total \\
        \midrule
        UR3   & 2,000 & 1,097 & 439 & 635 & 1,250 & 1,139 & 1,159 & 7,719 \\
        UR5   & 2,053 & 1,180 & 663 & 809 & 1,249 & 993 & 1,234 & 8,181 \\
        UR10  & 2,012 & 1,165 & 868 & 989 & 1,250 & 866 & 1,241 & 8,391 \\
        UR30  & 1,753 & 1,217 & 985 & 1,056 & 1,250 & 1,086 & 1,235 & 8,582 \\
        KR10  & 2,498 & 1,197 & -- & 1,103 & 1,248 & 1,176 & 1,649 & 8,871 \\
        \midrule
        Total & 10,316 & 5,856 & 2,955 & 4,592 & 6,247 & 5,260 & 6,518 & 41,744 \\
        \bottomrule
    \end{tabular}
\end{table}

\section{Per-Category Anomaly Detection on voraus-AD}
\label{app:per_category}
\Cref{tab:per_category_auroc} breaks the mean AUROC of
\cref{tab:anomaly_baselines} down by fault category. The S-E-F-C MLP is strongest
where the fault changes the load the controller has to hold, such as miscommutation
(99.2) and additional axis weight (95.8, the best of any method), and weakest
on brief gripping failures such as collision with cables (67.6) and losing the
can (71.8). Those are short, contact-driven events, and a single-step regressor
from Setpoint to Effort has no temporal context with which to see them. The
pattern is consistent with the role ablation in \cref{sec:sefc_ablation}: what
the model can detect is bounded by what the target role records.

\begin{table}[!ht]
    \centering
    \caption{Per-category anomaly detection AUROC on voraus-AD.
        Unstructured baselines use 130 signals; S-E-F-C MLP uses only 24 signals.
        $N$ = number of anomalous episodes per category.
        $^{\dagger}$\,95\% bootstrap CI over episodes, for the S-E-F-C MLP only. $^{\ddagger}$\,Baseline deviations are those reported by \citet{Brockmann2023TheVD} for each method and are not computed here; they are not comparable to the bootstrap interval, and the aggregate-row uncertainty in Table~\ref{tab:anomaly_baselines} is defined separately.}
    \label{tab:per_category_auroc}
    \resizebox{\textwidth}{!}{%
        \begin{tabular}{lrcccccccc}
            \toprule
            \textbf{Fault Category} & \textbf{$N$}
                                    & \textbf{1-NN}                           & \textbf{PCA}
                                    & \textbf{GANF}                           & \textbf{CAE}
                                    & \textbf{LSTM-VAE}                       & \textbf{HMM}
                                    & \textbf{MVT-Flow}
                                    & \textbf{S-E-F-C MLP (Ours)}                                                                                                                                                  \\
            \midrule
            Additional friction     & 144                                     & 74.8         & 76.4  & $88.5_{\pm4.5}$  & $89.4_{\pm0.2}$  & $88.7_{\pm1.5}$  & $88.0_{\pm0.7}$ & $96.6_{\pm0.6}$  & 80.0          \\
            Miscommutation          & 89                                      & 80.8         & 87.0  & $98.8_{\pm1.1}$  & $99.1_{\pm0.0}$  & $98.1_{\pm0.5}$  & $93.7_{\pm1.3}$ & $99.8_{\pm0.3}$  & 99.2          \\
            Misgrip can             & 11                                      & 100.0        & 100.0 & $47.6_{\pm13.1}$ & $100.0_{\pm0.0}$ & $100.0_{\pm0.0}$ & $71.0_{\pm3.3}$ & $95.3_{\pm3.3}$  & 83.8          \\
            Losing can              & 74                                      & 68.7         & 70.1  & $72.1_{\pm5.8}$  & $72.6_{\pm0.3}$  & $70.4_{\pm2.9}$  & $88.8_{\pm1.0}$ & $96.2_{\pm0.4}$  & 71.8          \\
            Add.\ axis weight       & 156                                     & 75.0         & 79.2  & $93.2_{\pm2.4}$  & $93.5_{\pm0.1}$  & $82.7_{\pm1.1}$  & $89.6_{\pm1.2}$ & $94.1_{\pm0.7}$  & \textbf{95.8} \\
            Collision w/ foam       & 72                                      & 69.6         & 73.9  & $81.2_{\pm6.2}$  & $81.5_{\pm0.2}$  & $81.5_{\pm2.1}$  & $89.8_{\pm1.5}$ & $87.5_{\pm1.2}$  & 74.7          \\
            Collision w/ cables     & 48                                      & 74.5         & 75.7  & $82.7_{\pm6.4}$  & $79.6_{\pm0.3}$  & $77.3_{\pm3.6}$  & $91.8_{\pm1.4}$ & $84.7_{\pm1.2}$  & 67.6          \\
            Collision w/ cardboard  & 22                                      & 82.8         & 83.6  & $77.5_{\pm5.8}$  & $78.6_{\pm0.3}$  & $82.8_{\pm4.8}$  & $86.3_{\pm1.2}$ & $88.3_{\pm1.2}$  & 76.7          \\
            Varying can weight      & 80                                      & 63.7         & 64.2  & $68.9_{\pm8.7}$  & $72.3_{\pm0.3}$  & $71.4_{\pm2.1}$  & $90.9_{\pm2.0}$ & $85.1_{\pm1.1}$  & 76.2          \\
            Cable at robot          & 10                                      & 63.0         & 71.6  & $76.6_{\pm8.1}$  & $83.1_{\pm0.3}$  & $96.0_{\pm1.4}$  & $84.4_{\pm1.1}$ & $100.0_{\pm0.0}$ & 91.3          \\
            Invalid gripping pos.   & 12                                      & 93.4         & 92.1  & $86.6_{\pm10.8}$ & $88.8_{\pm0.3}$  & $97.7_{\pm1.3}$  & $91.8_{\pm1.4}$ & $100.0_{\pm0.0}$ & 89.1          \\
            Unstable platform       & 37                                      & 83.6         & 85.9  & $84.6_{\pm3.8}$  & $83.9_{\pm0.2}$  & $93.6_{\pm1.9}$  & $82.5_{\pm1.2}$ & $96.1_{\pm0.7}$  & 92.8          \\
            \midrule
            \textbf{Mean}           & 755                                     & 77.5         & 80.0
                                    & $79.9_{\pm12.7}{}^\ddagger$
                                    & $85.2_{\pm9.2}{}^\ddagger$
                                    & $86.7_{\pm10.1}{}^\ddagger$
                                    & $87.4_{\pm5.8}{}^\ddagger$
                                    & $93.6_{\pm5.7}{}^\ddagger$
                                    & \textbf{83.2 [81.5,\,86.1]$^{\dagger}$}                                                                                                                                      \\
            \bottomrule
        \end{tabular}%
    }
\end{table}

\section{Simulator Physics and Controller}
\label{app:sim_physics}
\paragraph{Engine.} Isaac Sim 5.1.0 with Isaac Lab 2.3.2, GPU PhysX, headless,
batched (1,250 environments per wave). Physics steps at 500~Hz (64/64 solver
iterations); control and logging run every 4 physics steps (125~Hz) for the UR
arms and every 6 (83.3~Hz, the KRC interpolation cycle) for the KR10.

\paragraph{Cell and task.} Two stations with colliding decks, a ground plane and
a dynamic cuboid part. The UR3 cell follows our laboratory cell; the UR5, UR10
and UR30 cells are the UR3 cell scaled horizontally by the ratio of reach (1.88,
2.84, 2.88). The KR10 cell is fitted to real KR10 station angles. The task
shuttles the part between the stations (direction drawn per episode) in ten
phases: above pick, descend, settle, close, lift, move, lower, release, retreat,
return.

\paragraph{Motion layer.} A software model of the vendor motion layer generates
trapezoidal joint and Cartesian profiles with analytic inverse kinematics (all
eight branches, batched in torch) and a fixed configuration branch. The UR3
profiles are fitted to 109 real UR3 pick-and-place episodes (joint speed
0.96~rad/s, acceleration 1.03~rad/s$^2$; linear acceleration 0.54~m/s$^2$), the
KR10 profiles to 280 real KR10 nominal episodes (0.504~rad/s, 0.63~rad/s$^2$).
The UR5, UR10 and UR30 profiles have no real data: the UR3 profile is scaled
with the arm's reach and wrist, an assumption we declare rather than validate.

\paragraph{Joint control.} Each joint has an implicit PD drive (stiffness $k_p$
from Table~\ref{tab:sim_gains}, damping $2\zeta\sqrt{k_p(M_{ii}+I_a)}$ with
$\zeta=1$) plus computed-torque feedforward (gravity, Coriolis, inertia) clamped
to 0.8 of the effort limit and a joint-friction prior (UR3 Coulomb 2, 2, 1.5,
0.6, 0.6, 0.6~N\,m). The UR3 stiffnesses are set from a torque disturbance of
1.5~mrad; the UR5, UR10 and UR30 values are the UR3 values scaled by the ratio of
effort limits and were not measured. A protective-stop monitor is included. The
gripper is one actuated joint with five mimic joints; its stiffness is derived
each episode from the grip force needed to hold the sampled mass at the sampled
friction, clipped to $[2000, 40000]$.

\begin{table}[h]
    \centering
    \caption{Arm drive parameters (joints 1--6).}
    \label{tab:sim_gains}
    \footnotesize
    \setlength{\tabcolsep}{3pt}
    \begin{tabular}{lll}
        \toprule
        Arm & $k_p$ (N\,m/rad) & Effort limit (N\,m) \\
        \midrule
        UR3  & 6000, 6000, 4000, 800, 800, 600 & 56, 56, 28, 12, 12, 12 \\
        UR5  & 16071, 16071, 21429, 1867, 1867, 1400 & 150, 150, 150, 28, 28, 28 \\
        UR10 & 35357, 35357, 21429, 3733, 3733, 2800 & 330, 330, 150, 56, 56, 56 \\
        UR30 & 79071, 79071, 61857, 7133, 7133, 5350 & 738, 738, 433, 107, 107, 107 \\
        KR10 & 8000, 34000, 12000, 1900, 1900, 700 & 550, 650, 200, 35, 30, 20 \\
        \bottomrule
    \end{tabular}
\end{table}

\paragraph{Known deviations from the real robots.}
(i) The real UR and KRC controllers are proprietary; the model reproduces their
motion profiles, not their internal control law.
(ii) In the released corpus the commanded joint position is logged one control
tick ahead of the measured position, so the logged joint tracking error is
approximately $-8$~ms times the commanded velocity, which inflates nominal
tracking-error magnitude three to six times over the real robot. The calibrated
case study (Section~\ref{sec:transfer}) corrects the alignment; the main corpus
does not.
(iii) The payload feedforward is applied for the nominal part mass for the whole
episode with the wrong sign, adding the part's weight instead of compensating
it; a corrected path exists but was used only in the calibration runs.
(iv) The simulated TCP ``tracking error'' channel of an early ten-channel set is
the distance to the current pick or place target, not a control error; the
transfer study does not use it.
(v) The protective-stop monitor is effectively disarmed on the UR5 shoulder lift
and UR10 joints 1--4, and rare contact impulses occur on the UR10 shoulder lift.
(vi) Part dimensions are drawn once per process and reused across waves.

\section{Real-Data Transfer Summary}
\label{app:transfer_summary}
\Cref{tab:transfer_summary} collects the real-data transfer cells that do not
appear in the main-text figures.

\begin{table}[t]
    \centering
    \caption{Real-data transfer cells not shown in Figures~\ref{fig:transfer_phase}
    and~\ref{fig:transfer_fault}. Phase:
        macro-F1 (10 stages; ``5 stages'' = the five stages every real session
        emits). Faults: AUROC. Each real test set is one recording session;
        full cells in \cref{app:transfer}.}
    \label{tab:transfer_summary}
    \small
    \setlength{\tabcolsep}{4pt}%
    \begin{tabular}{llcc}
        \toprule
        Task & Train $\to$ test & UR3 target & KR10 target \\
        \midrule
        Phase & real, same session (ceiling) & 0.928 & 0.661 \\
              & real, same robot, other program/session (5 stages) & 0.78--0.80 & 0.69--0.72 \\
              & real, other robot & 0.227 & 0.145 \\
              & real, other robot (5 stages) & 0.395 & 0.318 \\
              & sim, other manufacturer & 0.377 & 0.366 \\
        \midrule
        Faults & real, same session (joint err.\ + current) & 0.85--0.90 & 0.90 \\
               & \quad speed-clean faults only (11, 30) & -- & 0.83 / 0.82 \\
               & real, other session, seen fault types & 0.64--1.00 & -- \\
               & real, other session, unseen fault types & 0.41--0.53 & -- \\
               & real, other robot (TCP) & 0.51--0.67 & 0.70$^{\ast}$ \\
        \bottomrule
    \end{tabular}\\[2pt]
    {\footnotesize $^{\ast}$Driven by speed-confounded faults; the speed-clean
    faults 11 and 30 score 0.45 and 0.67.}
\end{table}

\section{Batch sim-to-real validation: an earlier UR3 campaign}
\label{app:sim2real_gap}
This subsection reports an earlier UR3 validation campaign, run before the
five-arm corpus above. We run sim2real rollouts in Isaac Sim 4.5.0 (headless Docker, PhysX) using
the built-in UR3 USD asset and a position-control stack
(\texttt{ArticulationAction}) with phase-consistent waypoint replay from real
\texttt{target\_joint\_*} signals. The task is UR3 pick-and-place with a
10-phase structure (\texttt{above\_pick} to \texttt{return}). Simulation steps
at 240\,Hz and logs at 10\,Hz. Real episodes are exported from the laboratory recording stack to per-episode RTDE-shaped CSVs, converted to per-episode simulation
configs, replayed in Isaac, and paired by episode ID/filename. Simulated logs
preserve recording-stack-compatible fields (\texttt{joint\_*},
\texttt{target\_joint\_*}, \texttt{joint\_current\_*}, \texttt{tcp\_*},
\texttt{task\_phase}, and status/context fields). Gap metrics are computed
phase-wise with time normalization/interpolation. Results of the gap analysis are shown in Table~\ref{tab:sim2real_batch_gap}.

\begin{table}[h]
    \centering
    \caption{Batch sim-to-real gap over 1,155 paired episodes (pooled per-episode metrics).}
    \label{tab:sim2real_batch_gap}
    \small
        \begin{tabular}{lrrrr}
            \toprule
            Metric                 & Mean  & Median & P10   & P90   \\
            \midrule
            Joint RMSE (deg)       & 3.65  & 2.83   & 1.93  & 5.12  \\
            TCP position RMSE (mm) & 13.16 & 13.26  & 8.87  & 16.92 \\
            EE L2 RMS (mm)         & 25.11 & 25.17  & 16.96 & 32.58 \\
            TCP rotvec RMSE (mrad) & 1877  & 2605   & 8.75  & 3136  \\
            W1 effort mean (A)     & 0.83  & 0.81   & 0.77  & 0.91  \\
            \bottomrule
        \end{tabular}
\end{table}

An important source of residual pose discrepancy is end-effector mismatch: the
Isaac setup used a Robotiq 2FG85-style gripper configuration, whereas the real
laboratory episodes used an OnRobot 2FG14 gripper. Differences in tool
geometry/TCP definition and mounting can bias absolute TCP and orientation
metrics. We therefore interpret remaining TCP rotation spread conservatively
and treat gripper-accurate tool calibration as future work. Whether the
simulator is close enough to the real robots for a model to transfer is
tested directly in Section~\ref{sec:transfer}.

\section{Transfer study: protocol and supporting results}
\label{app:transfer}

\subsection{Protocol}
\label{app:transfer_protocol}

\paragraph{Data.} Simulated pick-and-place episodes on five arms (UR3, UR5, UR10,
UR30, KUKA KR10), each with the same ten-stage task program
(\texttt{above\_pick}, \texttt{descend}, \texttt{settle}, \texttt{close},
\texttt{lift}, \texttt{move\_xy}, \texttt{lower}, \texttt{release},
\texttt{retreat}, \texttt{return}). Real targets are our laboratory UR3 and KR10
pick-and-place recordings; a \emph{session} is one recording program and day on
the UR3 and one recording source and day on the KR10.

\paragraph{Channels.} Phase identification uses the six channels both real
robots log: TCP position and velocity in a normalised task-space frame. Joint
space transfers even worse between robots in simulation (UR3$\to$KR10 phase
macro-F1 0.177 in joint space vs 0.433 in task space; KR10$\to$UR3 0.455 vs
0.536). Fault detection additionally uses per-joint tracking error (setpoint
minus position) and motor current where both are available. Sim-to-sim cells use
the same six TCP channels as the real cells. An earlier ten-channel sim set is
not used: its simulated TCP ``tracking error'' was the distance to the current
pick or place target (the commanded TCP position takes two values per episode),
so it encoded phase and reach directly.

\paragraph{Models and metrics.} Windows of 64 samples with stride 32 and a
modal-label threshold of 0.6. MiniRocket + ridge
\citep{dempster2021minirocket} is the main model; a logistic regression on
summary statistics is the control. Phase is scored with macro-F1 over the ten
stages (union of true and predicted labels) or with balanced accuracy over the
seven stages that emit windows under this windowing (chance 0.143); every table
names its scale. Faults are scored with AUROC and a 95\% bootstrap interval,
grouped by counterfactual baseline where one exists. Shuffled-label nulls use
five permutations; we report their maximum. Five draws do not calibrate a null
distribution, so throughout we read these maxima as exploratory diagnostics and
not as significance thresholds; where a null run uses a different channel
configuration from the result it accompanies, we say so at that point.

\paragraph{Splits.} Real fault tests stay inside one two-class session and are
split into halves A and B. For UR3 fault 12 in the case study
(\cref{app:transfer_calib}) the halves hold 55/48 (A) and 54/51 (B)
nominal/fault episodes; for the KR10 session with in-session nominals, 21
nominals per half.

\subsection{Sim-to-sim transfer}
\Cref{tab:transfer_loo} holds out one simulated arm at a
time; \cref{tab:transfer_urmatrix} gives the pairwise matrix between single UR arms.
\label{app:transfer_sim}

\begin{table}[!ht]
    \centering
    \caption{Leave-one-arm-out transfer in simulation (six TCP channels,
        MiniRocket, trained on the other four arms; logistic-regression control
        in brackets; 95\% bootstrap interval below). ``Ceiling'' is the held-out
        arm trained on its own data. Test sets: 1,000 episodes (faults) and
        2,059--2,131 windows (phase) per arm.}
    \label{tab:transfer_loo}
    \small
    \resizebox{\linewidth}{!}{%
    \begin{tabular}{lccccccc}
        \toprule
        Held-out arm & \multicolumn{3}{c}{Fault AUROC} & \multicolumn{3}{c}{Phase, 7-stage bal.\ acc.} & Phase macro-F1 \\
        \cmidrule(lr){2-4}\cmidrule(lr){5-7}\cmidrule(lr){8-8}
        & LOO & ceiling & \% & LOO & ceiling & \% & LOO \\
        \midrule
        UR3  & 0.842 [0.720] & 0.947 & 89 & 0.994 [0.836] & 0.998 & 100 & 0.881 \\
             & {\scriptsize [0.814, 0.868]} & & & {\scriptsize [0.986, 0.999]} & & & \\
        UR5  & 0.913 [0.678] & 0.936 & 98 & 0.996 [0.829] & 0.997 & 100 & 0.919 \\
             & {\scriptsize [0.896, 0.930]} & & & {\scriptsize [0.992, 1.000]} & & & \\
        UR10 & 0.906 [0.665] & 0.940 & 96 & 0.986 [0.773] & 0.996 & 99 & 0.912 \\
             & {\scriptsize [0.878, 0.931]} & & & {\scriptsize [0.980, 0.993]} & & & \\
        UR30 & 0.939 [0.712] & 0.967 & 97 & 0.996 [0.780] & 0.992 & 100 & 0.929 \\
             & {\scriptsize [0.923, 0.952]} & & & {\scriptsize [0.994, 0.998]} & & & \\
        \midrule
        KR10 & 0.576 [0.580] & 0.776 & 74 & 0.812 [0.640] & 0.992 & 82 & 0.555 \\
             & {\scriptsize [0.539, 0.612]} & & & {\scriptsize [0.800, 0.824]} & & & \\
        \bottomrule
    \end{tabular}}
\end{table}

\begin{table}[!ht]
    \centering
    \caption{Transfer between single UR arms in simulation, six TCP channels
        (rows: trained on; columns: tested on). Fault AUROC / phase macro-F1.
        Reach: UR3 0.50\,m, UR5 0.85\,m, UR10 and UR30 1.30\,m. Pearson $r$
        between the 12 off-diagonal scores and $|\log$ reach ratio$|$: $-0.87$
        (faults), $-0.80$ (phase).}
    \label{tab:transfer_urmatrix}
    \small
    \begin{tabular}{lcccc}
        \toprule
        Trained on $\downarrow$ / tested on $\rightarrow$ & UR3 & UR5 & UR10 & UR30 \\
        \midrule
        UR3  & 0.947 / 0.995 & 0.764 / 0.897 & 0.605 / 0.680 & 0.602 / 0.689 \\
        UR5  & 0.782 / 0.880 & 0.936 / 0.970 & 0.825 / 0.865 & 0.835 / 0.855 \\
        UR10 & 0.776 / 0.767 & 0.849 / 0.777 & 0.940 / 0.800 & 0.948 / 0.915 \\
        UR30 & 0.554 / 0.719 & 0.731 / 0.875 & 0.887 / 0.848 & 0.967 / 0.940 \\
        \bottomrule
    \end{tabular}
\end{table}

\paragraph{Within the UR family transfer holds; across manufacturers it does
not.} A held-out UR arm reaches 99--100\% of its own phase ceiling (7-stage
balanced accuracy) and 89--98\% of its fault ceiling. A held-out KR10, trained on
the four UR arms, reaches 82\% of its phase ceiling (macro-F1 0.555) and 0.576
fault AUROC; that interval excludes 0.5 but lies within the range of our
shuffled-label runs (0.46--0.64; one permutation per held-out arm, on a
nine-channel variant). Pairwise, sim UR3$\to$KR10 scores 0.433 phase macro-F1
and KR10$\to$UR3 0.536 (within arm 0.995 / 0.984); fault AUROC 0.543 and 0.686.
The simulated arms share one task program, but the simulated KR10's nominal cycle
is about 1.6$\times$ longer than the UR3's (about 24 vs 15\,s), so simulation
does not separate embodiment from tempo.

\paragraph{Reach.} Among UR arms, transfer falls with reach distance: $r=-0.87$
(faults) and $-0.80$ (phase) over the 12 ordered pairs. Across 20 matched-budget
source pools, the reach distance of the pool's nearest member correlates with the
pool's score ($r=-0.765$ for faults, $-0.653$ for phase; farthest member $-0.274$,
$-0.230$). These analyses rest on four UR arms, so the 12 pairs are not
independent; that reach stands in for arm size and dynamics in general is a
hypothesis.

\paragraph{An honest simulated tracking error.} Adding a simulated TCP tracking
error computed by forward kinematics, $\mathrm{FK}(q_{\mathrm{cmd}}[k{+}1]) -
\mathrm{FK}(q[k])$, raises within-arm fault AUROC (UR3 0.947$\to$0.990, KR10
0.776$\to$0.900) but lowers cross-arm AUROC by 0.04--0.19, and removes the reach
relationship for faults ($r=-0.15$). A likely cause is that the error scale
differs between arms (a hypothesis). On its own this channel gives 0.19--0.53
leave-one-arm-out phase macro-F1, so it does not leak phase.

\subsection{Phase transfer between domains and robots}
\label{app:transfer_phase}
\Cref{tab:transfer_domain_grid} reports phase identification by training domain and
manufacturer at a matched budget, and \cref{tab:transfer_phasediag} varies one
processing choice at a time to diagnose the weak real cross-robot cell.

\begin{table}[!ht]
    \centering
    \caption{Phase identification by training domain and manufacturer at a
        matched budget ($\sim$2,445 windows, 6 channels). Macro-F1 / balanced
        accuracy (10-stage). Real tests are one session each; the real UR3 test
        session contains only 5 of the 10 stages, so compare within a column.
        Cells where the sim pool contains the target arm itself are within-arm
        references.}
    \label{tab:transfer_domain_grid}
    \small
    \begin{tabular}{lcccc}
        \toprule
        Trained on & $\to$ real UR3 & $\to$ real KR10 & $\to$ sim UR3 & $\to$ sim KR10 \\
        \midrule
        sim UR (4 arms)   & 0.317 / 0.570 & 0.366 / 0.451 & 0.934 / 0.905 & 0.531 / 0.624 \\
        sim KR10          & 0.377 / 0.594 & 0.351 / 0.434 & 0.614 / 0.648 & 0.992 / 0.991 \\
        sim UR + KR10     & 0.411 / 0.650 & 0.407 / 0.497 & -- & -- \\
        real UR3          & \textbf{0.928 / 0.929} & 0.145 / 0.222 & 0.238 / 0.316 & 0.248 / 0.320 \\
        real KR10         & 0.227 / 0.348 & \textbf{0.661 / 0.670} & 0.195 / 0.182 & 0.261 / 0.371 \\
        real UR3 + KR10   & 0.921 / 0.920 & 0.641 / 0.645 & -- & -- \\
        sim + real UR3    & 0.918 / 0.918 & 0.387 / 0.447 & -- & -- \\
        sim + real KR10   & 0.221 / 0.386 & 0.625 / 0.636 & -- & -- \\
        \bottomrule
    \end{tabular}
\end{table}

\begin{table}[!ht]
    \centering
    \caption{Diagnosing real cross-robot phase transfer. Each cell is macro-F1
        over the union of labels / macro-F1 retrained and tested on the five
        stages every real session emits (\texttt{above\_pick}, \texttt{close},
        \texttt{move\_xy}, \texttt{release}, \texttt{return}). Columns change one
        processing choice at a time.}
    \label{tab:transfer_phasediag}
    \small
    \resizebox{\linewidth}{!}{%
    \begin{tabular}{lccccc}
        \toprule
        Cell & Published & Derived vel., de-dup., tick clock & Tempo-free vel. & Raw time, 1.6\,s window & Per-episode z-score \\
        \midrule
        UR3$\to$KR10 & 0.145 / 0.318 & 0.225 / 0.343 & 0.280 / 0.322 & 0.180 / 0.337 & 0.400 / 0.402 \\
        KR10$\to$UR3 & 0.227 / 0.395 & 0.294 / 0.278 & 0.288 / 0.230 & 0.142 / 0.403 & 0.396 / 0.444 \\
        \midrule
        UR3 program A$\to$B / B$\to$A (5 stages) & 0.779 / 0.797 & \multicolumn{4}{l}{same robot, different recording program} \\
        KR10 session$\to$session (5 stages)      & 0.69--0.72    & \multicolumn{4}{l}{same robot, different session} \\
        \bottomrule
    \end{tabular}}
\end{table}

\paragraph{Label audit.} On both robots each labelled stage onset precedes its
physical event: \texttt{close} precedes the gripper closing by 0.08\,s (UR3) /
0.26\,s (KR10), \texttt{lift} and \texttt{retreat} precede the vertical rise by
0.14 / 0.44\,s, and \texttt{release} precedes the gripper opening by
0.16 / 0.46\,s; every lead is under 2\% of the cycle. The vertical-velocity
sign within each stage and the gripper width at each boundary match between the
robots. The task programs differ: median nominal cycle 16.0 vs 26.5\,s in the
test sessions; pick-to-place distance 0.43 vs 0.93\,m at the same vertical range (about
0.24\,m); descend and lift speed 0.10 vs 0.04\,m/s; a flat \texttt{move\_xy} on
the UR3 against an arc on the KR10; \texttt{return} takes 18.8\% of the UR3 cycle
and 7.7\% of the KR10 cycle. 
The UR3 test session contains
only five stages because \texttt{descend}, \texttt{settle}, \texttt{lift},
\texttt{lower} and \texttt{retreat} each last about 0.7\,s, too short for a
64-sample window; the KR10 test session contains nine. That sim sources beat
the other robot's real data may reflect that the four simulated UR arms span
several scales and tempos while one real UR3 program does not (a hypothesis;
untested on the five shared stages).

\paragraph{Few-shot (exploratory).} We also trained a CNN on simulation and
fine-tuned it on $k$ real UR3 episodes (6 channels, 7-stage balanced accuracy,
pooled over sessions; ceiling 0.922 with all 502 real episodes), and compared it
with training from scratch and with MiniRocket on pooled sim and real data. Each
point is the mean of three random draws of $k$ episodes, and every method sees the
same draws at a given $k$. The resulting curves are not monotone in $k$ (for
example, accuracy at $k=10$ is below $k=5$ for several methods), which indicates
that three draws do not average out episode-selection variance. We therefore do
not report a saving in real episodes as a result. The CNN reached 90\% of the
ceiling at about 352 real episodes from scratch and about 177 after pretraining,
where the curves are smoother; for MiniRocket the crossing lies on a noisy part of
the curve and we do not interpret it.

\subsection{Real-to-real fault detection by session}
\label{app:transfer_real}
\Cref{tab:transfer_sessions_counts} lists the recording sessions these tests draw on.

\begin{table}[!ht]
    \centering
    \caption{Real pick-and-place recording sessions (nominal / fault episodes,
        fault ids with counts where recorded, median cycle, log rate). Days are
        numbered from each robot's first recording day. Two-class sessions are
        the only ones that admit a within-session fault test. Counts are those of
        the session-fingerprint extraction; the full-release coverage count
        differs slightly for program 0010 days 1 and 2 (377/30 and 677/982
        episodes with signals).}
    \label{tab:transfer_sessions_counts}
    \small
    \resizebox{\linewidth}{!}{%
    \begin{tabular}{llllll}
        \toprule
        Robot & Session & Nom. / fault & Faults (id: $n$) & Cycle (s) & Log (Hz) \\
        \midrule
        UR3  & program 0010, day 1 & 345 / 29   & 19: 24, 37: 5 & 16.4 & 7.9 \\
        UR3  & program 0010, day 2 & 633 / 944  & 22, 25, 28 ($\sim$300 each), 19, 37 & 15.4 & 7.9 \\
        UR3  & program 0010, day 3 & 0 / 557    & 11, 22, 29, 30, 31, \ldots & 15.3 & 8.5 \\
        UR3  & program 0010, day 4 & 0 / 1072   & 23: 468, 15: 247, 10: 182, 38: 142 & 15.7 & 8.5 \\
        UR3  & program 0010, day 5 & 0 / 175    & 14: 175 & 14.8 & 8.5 \\
        UR3  & program 0010, day 6 & 0 / 321    & 8: 181, 9: 131 & 11.5 & 49 \\
        UR3  & program 0011, day 6 & 393 / 0    & -- (no grasp) & 5.3 & 49 \\
        UR3  & program 0012, day 6 & 109 / 327  & 11, 12, 29, 30 & 15.5--16.0 & 49 \\
        KR10 & source 1, day 1 & 19 / 0 & -- & 34.9 & 77.8 \\
        KR10 & source 2, day 2 & 205 / 0 & -- & 32.0 & 79.4 \\
        KR10 & source 2, day 3 & 0 / 353 & 12: 217, 38: 102, 28: 34 & 32.2 & 79.2 \\
        KR10 & source 4, day 4 & 0 / 242 & 28: 175, 11: 67 & 33.2 & 79.2 \\
        KR10 & source 3, day 6 & 44 / 0 & -- & 31.7 & 80.3 \\
        KR10 & source 4, day 6 & 20 / 214 & 25: 214 & 33.9 & 79.3 \\
        KR10 & source 5, day 7 & 42 / 288 & 10, 8, 11, 30 & 26.5 (nom.) & 80.1 \\
        \bottomrule
    \end{tabular}}
\end{table}

\Cref{tab:transfer_fulldata} compares the original and full UR3 releases.
The original analysis used one two-class UR3 pick-and-place session (program
0012). The full UR3 signal release raises this to three (program 0012 with
109/327 episodes and faults 11, 12, 29, 30; program 0010 day 1 with 377/30,
faults 19 and 37; program 0010 day 2 with 677/982, faults 19, 22, 25, 28, 37),
plus two two-class screwing sessions. Faults 8, 9, 10, 14, 15, 23, 31 and 38
remain without in-session nominals on the UR3. The KR10 data are unchanged.

\begin{table}[!ht]
    \centering
    \caption{Real fault detection before and after the full UR3 data
        (MiniRocket + ridge AUROC, mean of both halves, one session per test).}
    \label{tab:transfer_fulldata}
    \small
    \resizebox{\linewidth}{!}{%
    \begin{tabular}{lll}
        \toprule
        Analysis & Original data & Full data \\
        \midrule
        Within session, joint err.+current & 0.87 (program 0012) & 0012: 0.87; 0010 day 2: 0.95; 0010 day 1: 1.00 (29 faults) \\
        Within session, TCP only (6 ch.) & 0.65 (0012) & 0010 day 2: 0.70 (faults 25/28: 0.61; 19/22/37: 0.85--0.93); day 1: 1.00 \\
        Cross-session, same fault ids & not testable & faults 19/37: 0.64--1.00 per fault \\
        Cross-session, unseen fault ids & not testable & chance: 0012 0.41--0.53; 0010 day 2 faults 22/25/28 0.45--0.74 \\
        Phase, 7-stage bal.\ acc. & 0010: 0.937; 0012: 0.65 & 0010 day 2: 0.84 within; day 1: 0.66 within, 0.96 trained on day 2 \\
        \bottomrule
    \end{tabular}}
\end{table}

\paragraph{Pooled training.} \Cref{tab:transfer_pool} asks whether the small
within-session training sets limit the real fault ceilings. They do not:
pooling every other real episode of the same robot (regime b) moves the joint
tracking-error + current ceiling by $-0.10$ to $+0.04$, and adding the other
robot (regime c) costs up to a further 0.1. Most of the pooled episodes (77\% on
the UR3, 68\% on the KR10) come from single-class sessions; this suggests they
teach a program or session direction that the clean test penalises.

\begin{table}[!ht]
    \centering
    \caption{Fault AUROC with training regime a (within session: 216 UR3 / 166
        KR10 episodes) / b (all other episodes of the same robot: 930 / 1263) /
        c (b plus all episodes of the other robot). Tests: UR3 program 0012 and
        KR10 source 5 session, halves B and A. Shuffled-label null maximum
        $\le$0.61 (UR3), $\le$0.74 (KR10, TCP) and $\le$0.69 (KR10, joint).}
    \label{tab:transfer_pool}
    \small
    \resizebox{\linewidth}{!}{%
    \begin{tabular}{lcccc}
        \toprule
        Cell & UR3 half B & UR3 half A & KR10 half B & KR10 half A \\
        \midrule
        Joint err.+current, all phases (12 ch.) & 0.90 / 0.92 / 0.82 & 0.85 / 0.89 / 0.87 & 0.90 / 0.80 / 0.79 & 0.90 / 0.80 / 0.80 \\
        Current, all phases & 0.82 / 0.84 / 0.73 & 0.80 / 0.82 / 0.78 & 0.89 / 0.75 / 0.73 & 0.91 / 0.81 / 0.75 \\
        Tracking error, carry & 0.67 / 0.56 / 0.55 & 0.62 / 0.58 / 0.57 & 0.73 / 0.48 / 0.55 & 0.76 / 0.67 / 0.66 \\
        TCP, 6 ch. & 0.64 / 0.58 / 0.55 & 0.66 / 0.56 / 0.54 & 0.80 / 0.74 / 0.74 & 0.79 / 0.72 / 0.74 \\
        \bottomrule
    \end{tabular}}
\end{table}

\paragraph{KR10 speed confound.} In the KR10 source-5 session the fault episodes
ran at a lower speed override than the nominals. Speed alone scores AUROC 0.73
(half B) and 0.78 (half A); per fault, 0.83 (fault 8), 0.96 (fault 10), 0.61
(fault 11) and 0.57 (fault 30). With joint tracking error and current, the
speed-clean faults 11 and 30 score 0.83 and 0.82 under regime a and 0.58 and
0.73 under regime b.

\subsection{Cross-robot fault transfer}
\label{app:transfer_crossfault}

Real UR$\leftrightarrow$KR10 fault transfer stays at chance even with all data
pooled. With TCP channels, KR10$\to$UR3 scores 0.51--0.67, inside the
shuffled-label null ($\le$0.77); UR3$\to$KR10 scores 0.70, driven by the
speed-confounded faults 8 and 10, while the speed-clean faults 11 and 30 score
0.45 and 0.67. With joint channels, pooled cross-robot cells score 0.52--0.62,
at chance or inside the null (\cref{tab:transfer_uncal}). In a matched-budget domain grid, the training
domain (sim, real or both) explains 90\% of the variance of fault AUROC on real
targets and the manufacturer 0.4\%.

\subsection{Uncalibrated sim-to-real fault detection}
\Cref{tab:transfer_da} additionally asks whether feature-distribution alignment
rescues any of these cells; it does not.
\label{app:transfer_uncal}

\begin{table}[!ht]
    \centering
    \caption{Uncalibrated sim$\to$real fault AUROC (MiniRocket, 95\% bootstrap
        interval). Every interval contains 0.5. The 9-channel set adds TCP tracking
        error, which on the real UR3 is quantisation noise.}
    \label{tab:transfer_uncal}
    \small
    \begin{tabular}{llcl}
        \toprule
        Trained on (sim) & Tested on & Channels & AUROC \\
        \midrule
        KR10 + UR5 + UR10 + UR30 & real UR3 & 9 & 0.583 [0.487, 0.685] \\
        UR5 + UR10 + UR30 & real UR3 & 9 & 0.480 [0.381, 0.585] \\
        UR3 (same arm) & real UR3 & 9 & 0.491 [0.388, 0.597] \\
        KR10 & real UR3 & 9 & 0.493 [0.393, 0.603] \\
        same four pools & real UR3 & 6 & 0.473 / 0.460 / 0.459 / 0.443 \\
        UR3 + UR5 + UR10 + UR30 & real KR10 & 6 & 0.528 [0.457, 0.591] \\
        KR10 (same arm) & real KR10 & 6 & 0.504 [0.440, 0.563] \\
        \bottomrule
    \end{tabular}
\end{table}

\begin{table}[!ht]
    \centering
    \caption{Domain adaptation for uncalibrated sim$\to$real fault detection.}
    \label{tab:transfer_da}
    \small
    \begin{tabular}{lcl}
        \toprule
        Method & Fault AUROC before $\to$ after & Did the alignment itself work? \\
        \midrule
        Per-channel mean/sd matching & 0.493 $\to$ 0.553 & no (channels separable at 0.93--1.00) \\
        CORAL \citep{sun2016coral} & 0.501 $\to$ 0.519 & yes (domain classifier 1.000 $\to$ 0.521) \\
        DANN \citep{ganin2016dann} & 0.546 $\to$ 0.548 & no (encoder separable at 1.0) \\
        Task-space representation & 0.649 $\to$ 0.491 & n/a \\
        \bottomrule
    \end{tabular}
\end{table}

The reverse direction also fails: a real-trained fault model scores 0.335 on
simulated UR3. Pooling simulation with 200 real episodes scores lower than the
real episodes alone (0.583 vs 0.686); weighting both domains equally gives 0.681
vs 0.676 for real alone (not significant).

\subsection{Calibrated simulation: one UR3 payload fault}
\label{app:transfer_calib}

\Cref{tab:transfer_calib_components} lists what was calibrated and on which data, and
\cref{tab:transfer_calib} the resulting tracking-error scores.

\paragraph{Target fault.} In the real UR3 program 0012, fault 12 is a heavier box
(1.2\,kg in all fault episodes vs 0.6\,kg in nominals) with the declared payload
fixed at 1.5\,kg in both classes. On real data alone, carry-phase tracking error
detects it at 0.94 [0.91, 0.97] (grouped 5-fold; 0.89 in both half-to-half
directions; null maximum 0.65), and motor current at 0.99--1.00.

\begin{table}[!ht]
    \centering
    \caption{Calibration components of the simulator and the data each was
        fitted on. Only the timing-noise model was fitted strictly on the half
        not being scored.}
    \label{tab:transfer_calib_components}
    \small
    \resizebox{\linewidth}{!}{%
    \begin{tabular}{lll}
        \toprule
        Component & What it does & Fitted on \\
        \midrule
        Actual box masses & 0.6\,kg nominal, 1.2\,kg fault & real episode metadata (known design) \\
        Fault onset at grasp & mass switches at the first \texttt{lift} tick & \textbf{real fault-12 episodes} \\
        Declared-payload feedforward & controller compensates a fixed 1.5\,kg & real episode metadata \\
        Disturbance observer (load cancelling) & first-order, gain 1, $\tau$ 0.05--0.14\,s per joint & \textbf{real fault-12 episodes (both halves)} \\
        Command lag and acceleration term & lag 0.50--0.98\,ms, 101--266\,$\mu$s$^2$ & 109 real nominal episodes \\
        Place-height offset & +10\,mm at set-down & simulated episodes \\
        Aligned logging & removes a one-tick command/position offset & simulated twins, real nominals \\
        Logger emulation & replays the real logger's irregular sampling & real logger timing blocks \\
        Timing-noise model & 1-tick shifts + $\sim$0.4\,ms jitter + quantisation & real nominals of the \emph{other} half (strict) \\
        \bottomrule
    \end{tabular}}
\end{table}

\begin{table}[!ht]
    \centering
    \caption{Calibrated sim$\to$real UR3 fault 12, carry-phase tracking error
        (MiniRocket AUROC; real half B / half A / pooled). v2 and v3 differ in the
        number of simulated fault episodes (100 vs 400, each with a nominal twin).
        Noise arms are means over 5 noise seeds. Pooled 95\% intervals are about
        $\pm$0.06--0.08.}
    \label{tab:transfer_calib}
    \small
    \begin{tabular}{lcc}
        \toprule
        Simulator / logging model & v2 & v3 \\
        \midrule
        Uncalibrated simulator (half B) & \multicolumn{2}{c}{0.40 [0.29, 0.51]} \\
        Calibrated, regular logging & 0.66 / 0.69 / 0.67 & 0.58 / 0.60 / 0.59 \\
        + emulated real logger & 0.72 / 0.71 / 0.71 & 0.68 / 0.65 / 0.66 \\
        + white noise (strict) & 0.72 / 0.68 / 0.68 & 0.78 / 0.73 / 0.73 \\
        + timing-noise model (strict) & 0.78 / 0.73 / 0.76 & 0.82 / 0.77 / 0.79 [0.72, 0.85] \\
        \midrule
        Shuffled-label null maximum (half B) & \multicolumn{2}{c}{0.56--0.67} \\
        Real$\to$real ceiling (grouped 5-fold) & \multicolumn{2}{c}{0.94 [0.91, 0.97]} \\
        \bottomrule
    \end{tabular}
\end{table}

\paragraph{Motor current.} Across v2 and v3, regular and emulated logging, and
both halves, sim-trained current detection scores 0.90--0.99 AUROC (e.g.\ v3,
half B: carry 0.99 [0.97, 1.00], all phases 0.93 [0.88, 0.97]); the uncalibrated
simulator scores 0.62 (carry) and 0.41 (all phases). The logistic-regression
control on tracking error stays at 0.44--0.54 in every logging arm and 0.60 with
the timing-noise model, which suggests that the tracking-error result rests on
waveform shape rather than on signal level.

\paragraph{Caveats.} (i)~The disturbance observer and the logging diagnosis used
all real fault-12 episodes, so neither half is fully untouched; swapping the
halves is not an independent replication. (ii)~Gripper dwell alone separates real
fault from nominal episodes (AUROC 1.00); carry-phase cells exclude the dwell
phases, and within each class the carry score barely tracks dwell (Spearman
$+0.04$ nominal, $+0.11$ fault), but a class-linked recording confound cannot be
fully excluded. (iii)~The real$\to$real signal survives the controls we could
run: 0.87 on the 49 least dwell-mismatched pairs and 0.89 with nominals from the
fault recording window. (iv)~A load-aware linear model identified from real
nominal episodes only, used as a generator, reaches 0.90 on real tracking error;
this suggests the remaining sim gap lies in the load-transfer timing at lift-off
and set-down rather than in the classifier (a hypothesis).

\paragraph{KR10 check.} The only testable KR10 payload fault is fault 10 (an
undeclared +1.4\,kg between axes 4 and 5 for the whole episode; source-5 session,
42 nominals), and it is speed-confounded (speed alone 0.98). The UR3-calibrated
simulator scores 0.52 [0.41, 0.63] on carry tracking error and 0.42 [0.31, 0.53]
on carry current; a real UR3-trained model scores 0.12 and 0.41. The current
shift has the opposite sign on the two robots. A post-hoc, label-free sign
alignment on the three proximal joints gives 0.73 [0.64, 0.82]; because it was
chosen after seeing the data, in one speed-confounded session, we do not treat it
as confirmatory.

\subsection{Session fingerprints}
\label{app:transfer_sessions}
\Cref{tab:transfer_fingerprint} asks how well a session can be identified from the
signals alone, which bounds how much of any within-session result could be a
recording effect.

\begin{table}[!ht]
    \centering
    \caption{Session identification from signals only (grouped 5-fold, $\le$200
        episodes per session; balanced accuracy). \emph{Full}: per-phase
        mean/std of 21 channels plus phase durations. \emph{Timing}: phase
        durations and speed only. \emph{Shape}: MiniRocket on per-episode
        z-scored, phase-normalised signals (no level, scale, duration or speed).
        The label-shuffle null is at chance in every row.}
    \label{tab:transfer_fingerprint}
    \small
    \begin{tabular}{lcccc}
        \toprule
        Episodes (number of sessions) & Chance & Full & Timing & Shape \\
        \midrule
        UR3 nominal (4)  & 0.25 & 0.86 & 0.68 & 0.82 \\
        KR10 nominal (5) & 0.20 & 0.98 & 0.79 & 0.95 \\
        UR3 fault (7)    & 0.14 & 0.85 & 0.62 & 0.81 \\
        KR10 fault (4)   & 0.25 & 0.85 & 0.55 & 0.85 \\
        UR3, faults 19 and 37 only, 4 days of program 0010 & 0.25 & 0.65 & 0.64 & 0.61 \\
        \bottomrule
    \end{tabular}
\end{table}

Sessions are identifiable at 3--5$\times$ chance, and the signature survives the
removal of duration, speed and signal level. It lives in per-phase current and
tracking-error levels and in waveform shape rather than in any single
metadata-like variable; the exceptions are gross setup changes (a different
program, a 49\,Hz vs $\sim$8\,Hz log rate, a different speed override, a different
TCP configuration). Even the closest UR3 pair remains separable (AUROC
0.78--0.87 with all features). Any train/test split across sessions therefore
tests session transfer as well as fault detection. Group labels alone predict
the fault label: the UR3 recording program at 0.745 AUROC, the KR10 recording
source at 0.818 and calendar day at 0.937, and, within UR3 program 0010, the
recording day at 0.90 and the session at 0.91.

\end{document}